\pdfoutput=1

\documentclass[11pt]{article}

\usepackage[]{acl}

\usepackage{times}
\usepackage{latexsym}
\usepackage{xcolor}
\usepackage{makecell}
\usepackage{times}
\usepackage{latexsym}
\usepackage{hyperref}
\usepackage{inconsolata}
\usepackage{url}
\usepackage{makecell}
\usepackage{multirow}
\usepackage{graphicx}
\usepackage{caption}
\usepackage{subcaption}
\usepackage{amsmath}
\usepackage{bbm}
\usepackage{booktabs}
\usepackage{algorithmic}
\usepackage{amssymb}
\usepackage{bm}
\usepackage[ruled,vlined,linesnumbered]{algorithm2e}
\usepackage{enumitem}
\usepackage{pifont}
\usepackage{makecell}
\usepackage{pbox}

\usepackage{color}

\usepackage[T1]{fontenc}

\usepackage[utf8]{inputenc}

\usepackage{microtype}

\usepackage{inconsolata}

\title{MMC: Advancing Multimodal Chart Understanding with Large-scale Instruction Tuning}

\author{%
Fuxiao Liu$^{1,2}$\thanks{This work is done during internship at Tencent AI Lab.}, Xiaoyang Wang$^2$, Wenlin Yao$^2$, Jianshu Chen$^2$, \\ \textbf{Kaiqiang Song$^2$, Sangwoo Cho$^2$, Yaser Yacoob$^1$, Dong Yu$^2$}\\ 
$^1$University of Maryland, College Park $^2$Tencent AI Lab, Bellevue, USA\\ \small{\{fl3es, yaser\}@umd.edu,  \{shawnxywang, wenlinyao, jianshuchen, riversong, swcho, dyu\}@global.tencent.com}}

\begin{document}
\maketitle
\begin{abstract}
With the rapid development of large language models (LLMs) and their integration into large multimodal models (LMMs), there has been impressive progress in zero-shot completion of user-oriented vision-language tasks. However, a gap remains in the domain of chart image understanding due to the distinct abstract components in charts. To address this, we introduce a large-scale MultiModal Chart Instruction (\textbf{MMC-Instruction}) dataset comprising 600k instances supporting diverse tasks and chart types. Leveraging this data, we develop MultiModal Chart Assistant (\textbf{MMCA}), an LMM that achieves state-of-the-art performance on existing chart QA benchmarks. Recognizing the need for a comprehensive evaluation of LMM chart understanding, we also propose a MultiModal Chart Benchmark (\textbf{MMC-Benchmark}), a comprehensive human-annotated benchmark with nine distinct tasks evaluating reasoning capabilities over charts. Extensive experiments on MMC-Benchmark reveal the limitations of existing LMMs on correctly interpreting charts, even for the most recent GPT-4V model. Our work provides an instruction-tuning methodology and benchmark to advance multimodal understanding of charts.  Code and data are available at \url{https://github.com/FuxiaoLiu/MMC}.
\end{abstract}

\section{Introduction}
Large Language models (LLMs) such as GPT-3, PaLM, ChatGPT, Bard, and LLaMA \cite{brown2020language, chowdhery2022palm, openai2022ChatGPT, manyika2023overview, touvron2023llama, li2021character, xu2024survey} have undergone rapid development, demonstrating significant capabilities in performing a wide range of tasks effectively. 
To enable LLMs with vision ability, open-source large multimodal models (LMMs) such as MiniGPT-4~\cite{zhu2023minigpt}, LLaVA~\cite{liu2023visual}, mPLUG-Owl~\cite{ye2023mplug}, Multimodal-GPT~\cite{gong2023multimodal}, and LRV~\cite{liu2023aligning} have been developed, incorporating advanced image understanding capabilities into LLMs to interpret and analyze visual inputs. 
While successful in the general domains, such open-source LMMs are less effective for chart images because chart understanding differs tremendously from natural scene image understanding. In contrast with natural scene images, which primarily contain objects and reflect their spatial relationships, chart images contain unique abstract elements, including trend lines and color-coded legends that convey specific data-related information.

Current open-source LMMs are limited in their ability to accurately interpret complex chart contents, as they often lack domain-specific training essential for tasks such as differentiating between various types of graphs, interpreting axis labels and data points, and extracting meaningful patterns and trends. Integrating advanced chart understanding capabilities could further refine the LMMs' ability to analyze contextually and reason about the information presented in charts, thereby broadening their applicability in fields like data analytics, academic research, and business intelligence.

 \begin{figure*}[h]
    \centering
      \includegraphics[width=1\textwidth]{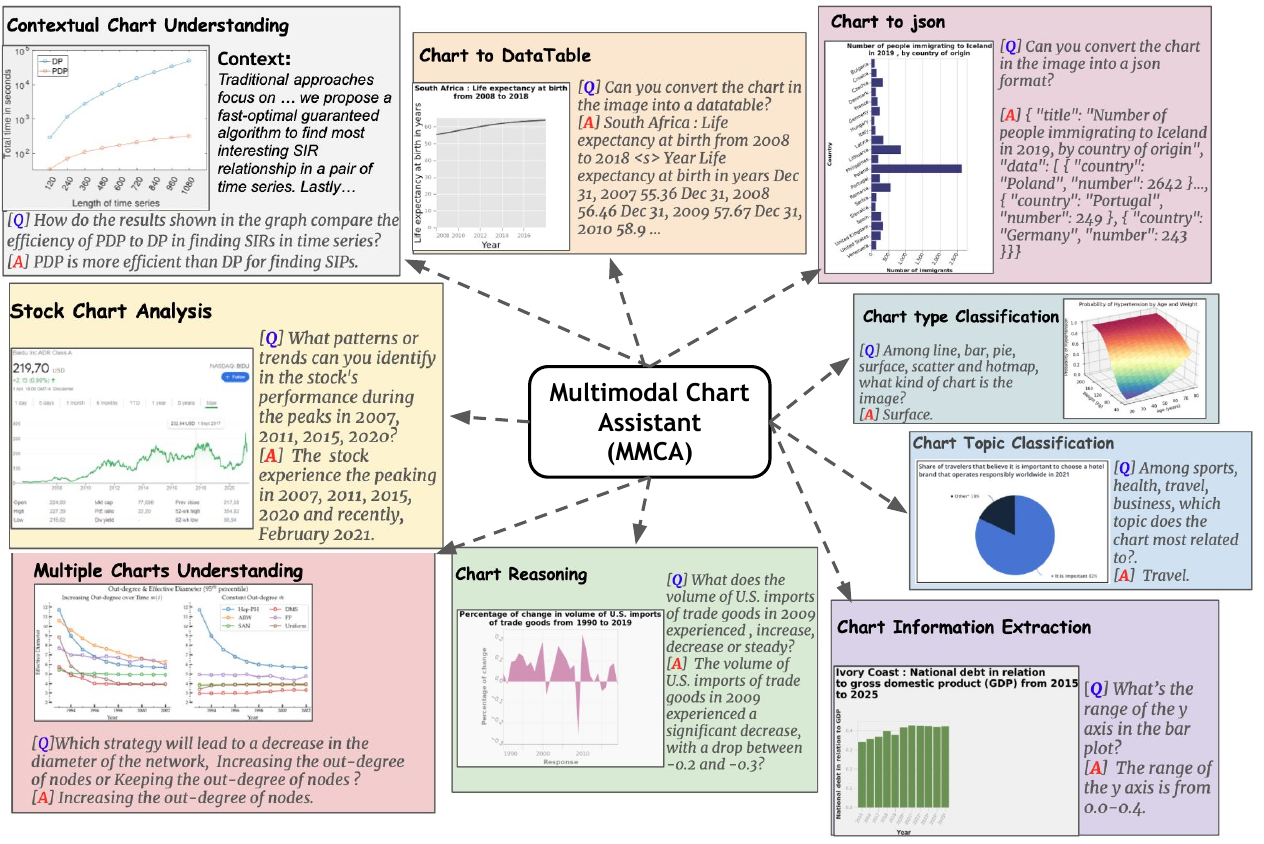}
    \caption{Diagram of our human-annotated \textit{MMC}, consisting of \textbf{nine} distinct tasks, various topics (\textbf{\textit{business, health, biology, engineering, etc}}), various chart types (\textbf{\textit{bar, histograms, line, scatter, heatmap, etc}}), free-form questions and open-ended answers. More examples are shown in the Appendix.}
    \label{fig:dataset_examples}
\end{figure*}

\begin{table*}[h]
\setlength\tabcolsep{4.3pt}
\centering
\small
\begin{tabular}{lccccccc}
\toprule[1pt]
\textbf{Datasets} & \textbf{Fig. Num} & \textbf{Question} & \textbf{Ans. Type} & \textbf{Ans. Length} &\textbf{Plot Type} & \textbf{Task Num}& \textbf{Benchmark} \\
\midrule
FigureQA &  180k &  Template &  Fixed Vocab &  1.0 & 4 &  1& \textcolor{black}{\ding{56}}\\
DVQA &  300k &  Template &  Fixed Vocab &  1.0 & 1 &  1& \textcolor{black}{\ding{56}}\\
PlotQA &  224k &  Template &  Fixed &  1.0 & 1&  1& \textcolor{black}{\ding{56}}\\
ChartQA &  21.9k &  Free-form &  Open Vocab &  1.2 & Unbounded&  2& Human Check\\
SciGraphQA &  295k &  Free-form &  Open Ended &  - &  Unbounded&  2&  \textcolor{black}{\ding{56}}\\
\midrule
\textbf{MMC-Instruction} &  600k &  Free-form &  Open Ended/MQA &  23.7 & Unbounded&  9 & Human Check\\

\bottomrule[1.5pt]
\end{tabular}
\vspace{0.05in}
\caption{Comparison between \textit{MMC-Instruction} with existing \textbf{chart} question-answering datasets. \textit{MQA} means multiple-choice question answering. \textit{MMC-Instruction} is \textbf{larger} and \textbf{more diverse}. ``Ans.'' stands for ``Answer''.}
\label{tab:dataset_statistic}
\vspace{-0.20in}
\end{table*}

In this paper, we introduce MultiModal Chart Instruction (\textbf{MMC-Instruction}), a 600k chart understanding dataset consisting of both chart-text alignment data and chart instruction-tuning data. \textit{MMC-Instruction} is not only much larger but also more diverse compared to existing public datasets \cite{kahou2017figureqa, masry2022chartqa, methani2020plotqa, kafle2018dvqa}.  Unlike previous work with templated-based questions, \textit{MMC-Instruction} is constructed by prompting GPT-4~\cite{openai2023GPT4} to generate instructions with diverse language styles and tasks (Tab.~\ref{tab:dataset_statistic}).
Furthermore, our \textit{MMC-Instruction} considers a variety of chart types, including but not limited to histograms, scatter plots, area charts, and more complex graphical representations. 
By performing unified instruction tuning upon current LMMs with \textit{MMC-Instruction}, we further propose a modularized LMM, namely Multimodal Chart Assistant (\textbf{MMCA}), jointly finetuned on a wide range of visually situated language understanding tasks. \textit{MMCA} achieves state-of-the-art performance on current chart question-answer benchmarks compared with existing open-source LMMs.

To accurately assess the capabilities of current Large Multimodal Models (LMMs) for chart understanding, we introduce a novel comprehensive evaluation tool: the MultiModal Chart Benchmark (\textbf{MMC-Benchmark}). First, \textit{MMC-Benchmark} is the first human-annotated benchmark in line with human cognition to evaluate LMM’s ability to comprehend visual charts. Second, it contains a wide range of tasks, including \textit{chart information extraction}, \textit{chart reasoning}, \textit{contextual chart understanding}, \textit{chart topic classification}, \textit{stock chart analysis}, \textit{multiple chart understanding}, \textit{chart type classification}, \textit{chart-to-datatable} and \textit{chart-to-json}. Third, \textit{MMC-Benchmark} offers two quantitative evaluation methods, including free-format Generation Ability Evaluation using GPT-4 and multiple-choice QA format Chart Understanding Ability Evaluation without the requirement of GPT-4.
Our evaluation highlights the limitations of existing open-source LMMs. In addition, we further broaden our analysis through experiments with GPT-4V \citep{2023GPT4VisionSC,yang2023dawn,liu2023hallusionbench}, the latest multimodal version of GPT-4. Our experiments indicate that \textit{MMC-Benchmark} also poses significant challenges to GPT-4V, especially in \textit{Chart to Datatable} and \textit{Chart to Json} tasks. It indicates the importance of \textit{MMC-Instruction} corpus and \textit{MMC-Benchmark} in advancing multimodal understanding.

Our main contributions are as follows:
\begin{itemize}[nosep]
    \item \textbf{MMC-Instruction} dataset. We present a novel large-scale instruction-tuning dataset for chart understanding. It includes diverse topics, language styles, chart types, and open-ended answers in line with human cognition. 
    \item \textbf{MMC-Benchmark}. We present a manually annotated benchmark specifically designed to assess the capability of LMMs in chart understanding across nine distinct sub-tasks to ensure a comprehensive evaluation.
    \item \textbf{MMCA} model. We propose an instruction-tuned LMM model that outperforms existing open-source state-of-the-art LMMs for chart understanding on both existing chart understanding benchmarks and our benchmark.
\end{itemize}


\label{sec:intro}

\section{Related Work}

\textbf{Multimodal Large Language Model.} Recently, Large Language Models (LLMs) have shown strong performances in zero-shot tasks across multiple domains. Recent studies explore using LLMs for multi-modal task completion. One direction \cite{wu2023visual, yang2023mm, yang2023gpt4tools} uses ChatGPT as the intermediary to choose the best tools or experts for visual interpretation according to user's inquiries. Another direction is end-to-end training \cite{zhu2023minigpt, liu2023visual, liu2023aligning, ye2023mplug, yin2023survey, wu2023next, zhang2023transfer, cao2023towards,zhai2023investigating} utilizing LLMs and visual encoders to create integrated models for multimodal tasks with inter-connected parameters to relate them. These existing approaches perform well on general visual and language tasks like image captioning and visual question answering with strong language skills. However, when it comes to chart understanding, they often fall short due to a lack of specific training to bridge the chart information with the textual content. Our work enhances chart understanding by introducing a novel chart visual instruction-tuning corpus and chart understanding model.

\textbf{Chart Text Understanding.}
Another line of research \cite{kantharaj2022chart, masry2023unichart, lee2023pix2struct} is to train a high-resolution image encoder on a large image-text pair corpus to learn text recognition during the pretraining stage. However, these models rely on specific finetuning on different downstream datasets and cannot achieve open-domain multi-task understanding like LLMs or LMMs do. Earlier datasets such as \cite{kahou2017figureqa, chaudhry2020leaf, methani2020plotqa, masry2023unichart, liu2020visualnews, liu2023documentclip} primarily rely on synthetic data, with template-generated questions and answers selected from a fixed vocabulary. More recently, ChartQA~\cite{masry2022chartqa} utilizes real-world, web-crawled charts to develop its visual question-answering datasets, supplemented by human annotators. However, it mainly focuses on compositional and visual questions. \cite{li2023scigraphqa} uses Palm-2 to generate question-answering data for academic charts. However, the answers generated by Palm-2 contain hallucinations. Comparatively, the advantages of our dataset come from its larger size,  more diverse topics, richer language styles, and good quality.

\label{sec:related_work}

\section{MMC-Instruction}
\begin{table*}[t]
\begin{subtable}[t]{0.35\textwidth}
\small
\begin{tabular}{lcccc}
\toprule[1pt]
\textbf{Benchmark} & \textbf{Size} & \textbf{Images} & \textbf{Source}& \textbf{Answer} \\
\midrule
VQA &  >1M &  General &  Annotated &  Open\\
GQA &  >1M &  General &  Synthesized &  Open\\
MME &  1.5k &  General &  Annotated &  Y/N\\
Lynx-Bench &  0.5k &  Video &  Annotated &  Open\\
MMBench &  3k &  General &  Repurposed &  MQA\\
MM-Vet &  0.2k &  General &  Repurposed &  MQA\\
MathVista &  1.4k &  Math &  Synthesized &  MQA\\
\midrule
\textbf{MMC-Benchmark} &  2k &  Chart/Plot &  Internet, Annotated &  Open/MQA\\
\bottomrule[1.5pt]
\end{tabular}
\end{subtable}
\hspace{1.8in}
\begin{subtable}[t]{0.4\textwidth}
\setlength\tabcolsep{4.7pt}
\centering
\small
\begin{tabular}{lc}
\toprule[1.5pt]
Statistic & Num \\
\midrule
\textbf{\textit{MMC-Instruction}} & 600k\\
\textit{-- Scientific Chart-Caption} & 210k\\
\textit{-- Filtered Existing Datasets} & 190k\\
\textit{-- GPT-Generated Instructions} & 200k\\
\textbf{\textit{MMC-Benchmark}} & 2k\\
\textit{-- Unique number of images} &1,063\\
\textit{-- Multiple-choice questions} & 1,275\\
\textit{-- Free-form questions} & 851\\
\textit{-- Average question length} & 15.6\\
\bottomrule[1.5pt]
\end{tabular}
\end{subtable}
\vspace{-2mm}
\caption{Comparison between \textit{MMC-Benchmark} with existing vision-language benchmarks. \textit{MQA} means multiple-choice question answering. \textit{Repurposed} means the benchmark is a compilation of prior datasets. \textit{Y/N} means yes/no questions. \textit{MMC-Benchmark} is the only existing benchmark with high-quality images for chart understanding.}
\label{tab:benchmark_comparison}
\end{table*}

\begin{table*}[h]
\setlength\tabcolsep{4.3pt}
\centering
\small
\begin{tabular}{l|ccccc}
\toprule[1pt]
\textbf{Tasks} & \textbf{Image Source} & \textbf{Question Source} &\textbf{Question Type} & \textbf{Number}&Human Check \\
\midrule
Chart Information Extraction &  Statista.com &  GPT-4 & Free-form/MQA & 330 &\textcolor{black}{\ding{52}}\\
Chart Reasoning &  Statista.com &  GPT-4 & Free-form/MQA &256&\textcolor{black}{\ding{52}}\\
Contextual Chart Understanding &  arxiv &  GPT-4, human &Free-form/MQA & 56&\textcolor{black}{\ding{52}}\\
Multiple Chart Understanding &  arxiv &  GPT-4, human &Free-form/MQA & 52&\textcolor{black}{\ding{52}}\\
Chart Type Classification &  Web Crawl &  Groundtruth label &Free-form/MQA & 360&\textcolor{black}{\ding{52}}\\
Chart Topic Classification &  Web Crawl &  Groundtruth label &Free-form/MQA & 536&\textcolor{black}{\ding{52}}\\
Chart To DataTable &  VisText &  Source Article &Free-form/MQA & 400&\textcolor{black}{\ding{52}}\\
Chart To Json &  VisText &  GPT-4 & Free-form/MQA &96&\textcolor{black}{\ding{52}}\\
Stock Chart Analysis &  Google Bard &  Source Article &Free-form/MQA & 40&\textcolor{black}{\ding{52}}\\
\bottomrule[1.5pt]
\end{tabular}
\vspace{0.05in}
\caption{Compositions of \textit{MMC-Benchmark}. The distributions of topics and types are shown in Fig.~\ref{fig:type_stat} and Fig.~\ref{fig:topic_stat}.}
\label{tab:benchmark_statistic}
\vspace{-0.20in}
\end{table*}

\subsection{Chart-Text Alignment Data}
To build a large training corpus for chart-text alignment with a diverse range of styles and topics, we aim to collect chart and text data from online sources. We first collect the \textit{Scientific Chart-Caption} corpus with both chart and text crawled from arXiv. In addition, we filter several existing public datasets that are suitable for chart-text alignment. The collected charts can be categorized into multiple topics, including (\textit{computer science, business, health, biology, agriculture, etc.}), and a variety of chart types, including but not limited to \textit{(histograms, scatter plots, area charts, and heatmap)}. More statistic is shown in Tab.~\ref{tab:dataset_statistic} and Tab.~\ref{tab:benchmark_comparison}.

\textit{\textbf{Scientific Chart-Caption}} data collected by us. We first download the academic articles (2010-2020) through an official dump from the arXiv website. It is licensed under CC-0, which grants remake and republish rights. Unlike \cite{hsu2021scicap} using PDFs, we utilize the source files containing the original LaTeX and figure files. In order to improve the dataset quality, we removed the source files without LaTeX or figure files and the source files that are hard to parse. We only keep the chart figures with rich text information by deleting the pairs whose caption length is less than 25 tokens. Finally, we collect 210k chart-text pairs in total.

\textit{\textbf{Leveraging Existing Datasets.}} For chart-text alignment training with diverse chart caption data, we further include the following five public chart datasets for which the underlying data tables are available: (i) Statista~\cite{kantharaj2022chart}, (ii) PlotQA~\cite{methani2020plotqa},  (iii) VisText~\cite{tang2023vistext}, (iv) ChartInfo~\cite{lal2023lineformer}, (v) Unichart~\cite{masry2023unichart}. We randomly picked approximately 190k image-text pairs from these public datasets to increase the diversity.

\subsection{Chart Instruction-Tuning Data}
This section introduces the construction of our instruction tuning data with 200k instances. To align the model to follow a variety of instructions, we construct diverse instruction-tuning instances about the provided chart images by prompting the language-only GPT-4~\cite{openai2023GPT4}. Specifically, given a chart description, we design instructions in a prompt that asks GPT-4 to generate questions and answers in a style as if it could see the image (even though it only has access to the text). The prompt examples for GPT-4 are shown in Fig.~\ref{fig:chart_prompt_ie}, \ref{fig:chart_prompt_cr}, \ref{fig:chart_prompt_arxiv}, \ref{fig:chart_prompt_json}. Our instruction-tuning format is: ``Human: \{question\} AI: \{answer\}''. \textit{MMC-Instruction} includes the following tasks: \textit{chart information extraction}, \textit{chart reasoning}, \textit{scientific chart understanding}, \textit{chart-to-datatable}, and \textit{chart-to-json}.

\textbf{\textit{Chart Information Extraction}} requires the model to extract from the input chart detailed information such as title, coordinate value, scope, etc. To achieve this goal, we collect the generated L1 captions from \cite{tang2023vistext}, whose content enumerates aspects of the chart’s construction. Then, we ask GPT-4 to generate question-answer pairs about the detailed construction information about the chart given descriptions (Fig.~\ref{fig:chart_prompt_ie}). Additionally, we require the generated answers to be less than 20 words to address hallucination.

\textbf{\textit{Chart Reasoning}} requires the model to analyze and identify data patterns, relationships, and anomalies of the input chart. To achieve this goal, we collect the generated L2/L3 captions from \cite{tang2023vistext}, which summarize the statistics and synthesize the cognitive phenomena of the chart. Then, we ask GPT-4 to generate question-answer pairs that require analysis skills in Fig.~\ref{fig:chart_prompt_cr}.

\textbf{\textit{Scientific Chart Understanding}} is a challenging task that needs scientific background knowledge. To create instruction-tuning data, we combine the abstract, title, and image captions of arXiv papers to construct the comprehensive textual context. Sometimes, the image caption is too short for GPT-4 to generate meaningful questions and answers regarding the image. To provide more context regarding the image, we also created a prompt that included paragraphs mentioning the figure in the paper. From our observation, we find a portion of the questions are not graph-related but a follow-up on the textual context in previous answers. We use heuristic rules to delete the non-chart-related questions. The prompt is shown in Fig.~\ref{fig:chart_prompt_arxiv}.

\textbf{\textit{Chart-to-DataTable}} and \textbf{\textit{Chart-to-Json}} are the tasks of transforming the visual information represented in the chart into the structured data format of a table or a JSON. This process typically requires interpreting the graphical elements of the chart, such as bars, lines, or pie segments, quantifying their values, and then organizing these values into a tabular format that accurately reflects the original chart. As shown in Fig.~\ref{fig:chart_prompt_json}, we transform the groundtruth data table from \cite{tang2023vistext} to create the JSON format into our \textit{MMC-Instruction}.

\textbf{\textit{Further Quality Control.}} We first remove instances with answers longer than 20 words. We remove the instances mentioning unneeded content like  "given caption" and "existing descriptions". As for the \textit{Chart-to-Json} task, we remove the instances without mentioning \textit{"title"} as the key. To examine the quality of our dataset, we randomly sample 500 instances and ask expert annotators to determine whether the output answers from GPT-4 are correct or not, with regard to the instruction and the image content. We find that 91\% of the instructions are appropriate for the image inputs. Furthermore, 85\% of outputs are acceptable responses to the instructions. Though some responses may contain errors, most generations conform to the correct structure, serving as applicable instruction-tuning guidelines.

\section{MMC-Benchmark}
The recent progress of LMMs has enabled the open-ended zero-shot completion of user-oriented vision-language tasks such as open-ended chart understanding. As a result, a comprehensive evaluation benchmark is necessary to evaluate the performances of different LMMs on these tasks and provide quantitative guidance for future research and development. However, for chart understanding, existing benchmarks often fall short of evaluating open-ended questions and unbounded chart types. Our dataset, \textit{MMC-Benchmark}, is therefore motivated to bridge this gap, offering three unique characteristics for chart understanding:

(i) \textit{MMC-Benchmark} is the first benchmark with human annotations to evaluate LMM's ability to comprehend visual charts.

(ii) \textit{MMC-Benchmark} is more diverse with various sources and nine different tasks, including \textit{chart information extraction}, \textit{chart reasoning}, \textit{contextual chart understanding}, \textit{multiple chart understanding}, \textit{chart type classification}, \textit{chart topic classification}, \textit{chart-to-datatable}, \textit{chart-to-json}, and \textit{stock chart analysis}, with examples shown in Fig.~\ref{fig:dataset_examples}.

(iii) \textit{MMC-Benchmark} provides two evaluation methods for convenient quantitative analysis, including free-format Generation Ability Evaluation using GPT-4 and multiple-choice QA format Chart Understanding Ability Evaluation without the requirement of GPT-4. The statistic of \textit{MMC-Benchmark} is shown in Tab.~\ref{tab:benchmark_comparison} and Tab.~\ref{tab:benchmark_statistic}.
\vspace{-0.10in}
\subsection{Data Annotation and Quality Control} 
For \textit{chart information extraction} and \textit{chart reasoning} tasks, the images are samples from \cite{masry2022chartqa}, but the instruction-answer pairs are all manually constructed by us rather than from existing public annotations. For \textit{contextual chart understanding} and \textit{multiple chart understanding}, we collect the source images from scientific charts of arXiv that are not presented in our training sets. Contextual chart understanding requires the models to read the context information to answer the questions of the charts. We utilize the abstract of the scientific paper as the context information. We manually design the questions for the multiple chart understanding, which evaluates the model's complex reasoning ability to compare between multiple charts. The \textit{chart type classification} task contains seven types: line, bar, pie, scatter, heatmap, histogram, and Radar. The images of line, bar, and pie chart are from \cite{methani2020plotqa} while others are collected by us from Google Bard. The \textit{chart topic classification} task includes health, business, science, travel, biology, engineering, and sports, whose images are crawled from Google. As for the \textit{chart-to-datatable} and \textit{chart-to-json} tasks, we use the images and data tables from \cite{wu2023visual}. The json data is generated by prompting GPT-4 with the datatable as the input. Finally, for \textit{stock chart analysis}, we collect the chart images of stock from Google Bard without including corresponding captions due to hallucination concerns. Instead, we look through the source article and manually construct the questions about the stock trend, predictions, and corresponding background knowledge. We adhere to copyright and license regulations, avoiding data from sites prohibiting copy and redistribution. More examples are shown in Fig.~\ref{fig:dataset_examples},  \ref{fig:type_example}, \ref{fig:IE_example}, \ref{fig:multiple_example}, \ref{fig:topic_example}, \ref{fig:arxiv_example}, \ref{fig:stock_example}, \ref{fig:CR_example}, \ref{fig:json_example}. The topic and type distributions are shown in Fig.~\ref{fig:topic_stat} and \ref{fig:type_stat}.

\vspace{-0.10in}
\subsection{Evaluation Protocols} 
In order to evaluate LMMs' generation ability and chart understanding ability, the instructions in \textit{MMC-Benchmark} consist of two parts. 

\textbf{Generation Ability Evaluation} utilizes GPT-4 ``gpt-4-32k-0314'' to assess the accuracy of prediction given question and reference answers using prompts shown in Fig.~\ref{fig:prompt_chatgpt}. We randomly select 300 samples from our testing set and manually evaluate the model predictions. We find GPT-4 assisted evaluation can achieve 0.90 agreement (Cohen’s kappa agreement) with human evaluation.

\textbf{Understanding Ability Evaluation (MQA)}, which aims to let the model select the correct answer from multiple-choice questions (MQA) given the chart. For each image, we manually design choices for each question. \textit{Understanding Ability Evaluation} does not require the utilization of GPT-4. We adopt micro-averaged accuracy as the evaluation metric in ~\cite{yu2023mm} with the help of systematic, rule-based evaluation pipelines.

\label{sec:dataset}

\section{MultiModal Chart Assistant (MMCA)}
\begin{figure}[h]
    \centering
      \includegraphics[width=0.47\textwidth]{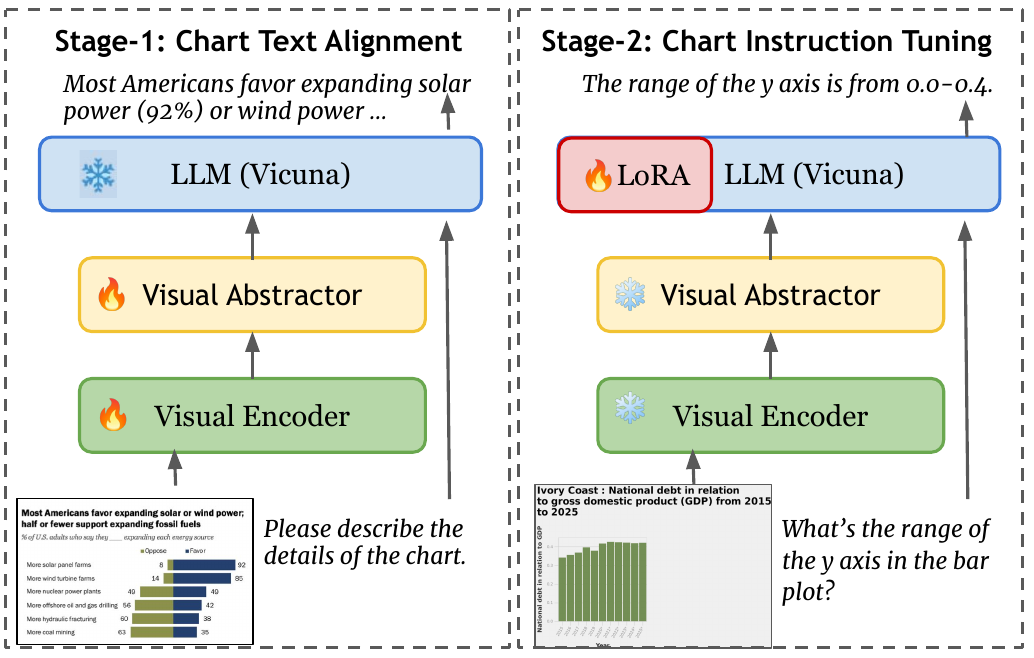}
    \caption{The overall architecture of \textsc{MMCA}, which is continuously trained in two stages.}
    \label{fig:model}
    \vspace{-0.2in}
\end{figure}

\textbf{Architecture.} Our model \textit{MMCA} (Fig.~\ref{fig:model}) is built on mPLUG-Owl~\cite{ye2023mplug}) that guides LLMs to follow multimodal instructions. In order to improve the existing LLMs to perform better on chart understanding tasks, we further fine-tune mPLUG-Owl 7B~\cite{ye2023mplug}) on our proposed \textit{MMC-Instruction} corpus consisting of \textit{Chart-Text Alignment Data} and \textit{Chart Instruction-Tuning Data}. mPLUG-Owl contains a pre-trained visual foundation model (CLIP vision encoder), a visual abstractor, and a language foundation model (Vicuna). The visual foundation model is responsible for extracting visual features from the input images, and the visual abstractor distills these features using a set of learnable tokens. The resulting visual features are combined with the word embeddings of the input sentence and fed into the language model to generate the response. We incorporate a two-stage training paradigm.

\textbf{Stage-1: Chart Text Alignment.} In this stage, we freeze the language decoder and train the visual parts with our \textit{Chart-Text Alignment Data} for one epoch. This stage enables the mapping of visual features of charts to LLM's word embedding space.

\textbf{Stage-2: Chart Instruction Tuning.} In the second stage, we freeze the visual abstractor, visual encoder, and language decoder and use the low-rank adaptation approach (LoRA)~\cite{ye2023mplug} to fine-tune the language model. Specifically, we train the language model with LoRA on our \textit{Chart Instruction-Tuning Data} for three epochs. This stage enables LLM's instruction following capabilities for chart understanding.

\label{sec:method}

\vspace{-0.10in}
\section{Experiments}

\subsection{Experimental Setup}
Our model training and inference are conducted with Tesla V100 GPUs. The evaluation is conducted under a zero-shot setting. More implementation details are discussed in the Appendix.

\vspace{-0.10in}
\subsection{Baselines}
We compare \textit{MMCA} with existing models in three setups: (a) Open-source LMMs including MiniGPT-v2-7B~\citep{chen2023minigpt}, mPLUG-owl-7B~\citep{ye2023mplug}, LRV-Instruction-7B~\citep{liu2023aligning}, LLaVA1.5-7B~\citep{liu2023improved}, and Multimodal-GPT-9B~\citep{gong2023multimodal}. (b) GPT-4V~\cite{2023GPT4VisionSC} by OpenAI. (c) Non-LLMs based models including Pix2Struct~\citep{lee2023pix2struct} and Donut~\citep{kim2022ocr}.

\subsection{Experiment Results}

\begin{table*}[t]
\setlength\tabcolsep{4.3pt}
\centering
\small
\begin{tabular}{l|ccccc|c}
\toprule[1pt]
\textbf{Free-form Evaluation} & \textbf{LLAVA1.5} & \textbf{MiniGPT-v2} & \textbf{mPLUG-Owl} & \textbf{LRV-Instruct}& \textbf{MMCA (Ours)}& \textbf{GPT-4V} \\
\midrule
Chart Information Extraction &  0.32 &  0.29 &  0.27 & 0.24 & \textbf{0.35} & \textbf{0.63}\\
Chart Reasoning &  0.30&  0.23 &  0.22 & 0.19 & \textbf{0.30}& \textbf{0.57}\\
Contextual Chart Understanding &  0.33 &  0.29 &  0.28 & 0.23 & \textbf{0.33} & \textbf{0.55}\\
Multiple Chart Understanding&  0.27 &  0.20 &  0.23 & 0.21 & \textbf{0.29} & \textbf{0.39}\\
Chart Type Classification &  0.30 &  0.27 &  0.25 & 0.22 & \textbf{0.31} & \textbf{0.79}\\
Chart Topic Classification &  0.31&  0.23 &  0.24 & 0.21 & \textbf{0.32}& \textbf{0.82}\\
Stock Chart Analysis &  0.27 &  0.28 &  0.25 & 0.23 & \textbf{0.32}& \textbf{0.70}\\
Chart to Datatable &  0.00&  0.00 &  0.05 & 0.00 & \textbf{0.08}& \textbf{\underline{0.05}}\\
Chart to Json  &  0.01&  0.00 &  0.00 & 0.00 & \textbf{0.05}& \textbf{\underline{0.04}}\\
\midrule
Overall &  0.24&  0.21 &  0.20 & 0.17 & \textbf{0.26}& \textbf{\underline{0.51}}\\
\bottomrule[1.5pt]
\end{tabular}
\vspace{0.05in}
\caption{\textit{MMC-Benchmark} evaluation results of LLaVA1.5, MiniGPT-v2, mPLUG-Owl, LRC-Instruct, MMCA, and the recent GPT-4V regarding the Generation Ability Evaluation. Given the reference response, we apply \textit{GPT-4} to determine the correctness/incorrectness (as in Fig.~\ref{fig:prompt_chatgpt}) of the response for each test sample. The ratio of correct responses out of responses for all test samples in each task is used for evaluation. Tab.~\ref{tab:baseline_size} shows the sizes of models.}
\label{tab:comparison_mmc}
\vspace{-0.1in}
\end{table*}

\begin{table*}[t]
\setlength\tabcolsep{4.3pt}
\centering
\small
\begin{tabular}{l|ccccc|c}
\toprule[1pt]
\textbf{MQA Evaluation} & \textbf{LLAVA1.5} & \textbf{MiniGPT-v2} & \textbf{mPLUG-Owl} & \textbf{LRV-Instruct}& \textbf{MMCA (Ours)}& \textbf{GPT-4V} \\
\midrule
Chart Information Extraction &  0.47 &  0.43 &  0.45 & 0.45 & \textbf{0.49}& \textbf{0.76}\\
Chart Reasoning &  0.45&  0.39 &  0.41 & 0.41 & \textbf{0.47}& \textbf{0.74}\\
Contextual Chart Understanding &  0.49 &  0.51 &  0.50 & 0.42 & \textbf{0.55}& \textbf{0.79}\\
Multiple Chart Understanding&  0.42 &  0.41 &  0.43 & 0.45 & \textbf{0.47}& \textbf{0.65}\\
Chart Type Classification &  0.55 &  0.52 &  0.55 & 0.50 & \textbf{0.59}& \textbf{0.85}\\
Chart Topic Classification &  0.59&  0.56 &  0.54 & 0.51 & \textbf{0.64}& \textbf{0.87}\\
Stock Chart Analysis &  0.52 &  0.49 &  0.45 & 0.45 & \textbf{0.57}& \textbf{0.81}\\
Chart to Datatable &  0.57&  0.46 &  0.44 & 0.35 & \textbf{0.64}& \textbf{0.71}\\
Chart to Json  &  0.51&  0.44 &  0.41 & 0.39 & \textbf{0.59}& \textbf{0.69}\\
\midrule
Overall &  0.51&  0.47 &  0.45 & 0.43 & \textbf{0.56}& \textbf{\underline{0.76}}\\
\bottomrule[1.5pt]
\end{tabular}
\vspace{0.05in}
\caption{\textit{MMC-Benchmark} evaluation results of LLaVA1.5, MiniGPT-v2, mPLUG-Owl, LRC-Instruct, MMCA and the recnet GPT-4V regarding the Understanding Ability Evaluation via \textit{Multichoice QA} (MQA) task. We calculate the accuracy of the model predictions in the MQA setting. There is no need to call \textit{GPT-4} for this evaluation.}
\vspace{-0.2in}
\label{tab:comparison_mmc_mvqa}
\end{table*}

\subsubsection{Evaluation Results on \textit{MMC-Benchmark}}
As indicated in Tab.~\ref{tab:comparison_mmc}, Tab.~\ref{tab:comparison_mmc_mvqa} and Tab.~\ref{tab:comparison_mmc_mvqa_more}, \textit{MMCA} achieves better performance in all nine tasks in comparison with the existing open-source models. The improvement of \textit{MMCA} demonstrates the effectiveness of our \textit{MMC-Instruction} data in enabling the LMM to complete chart understanding tasks. In addition, we find that current LMMs are better at understanding cross-modality relationships in the image but weaker at comprehending text layout information. This can be attributed to their lack of text recognition, scientific knowledge, and math reasoning abilities. Though fine-tuned with instruction-tuning data from text-rich images, LLAVa1.5 and mPLUG-Owl do not perform well, indicating that strong text recognition abilities in images do not guarantee high performance on \textit{MMC-Benchmark}, which requires comprehensive visual perception and chart reasoning capability. Additionally, current LMMs perform badly on the \textit{chart-to-datatable} and \textit{chart-to-json} tasks. We speculate it is because these two tasks require strong OCR ability to output all the data values in the chart correctly. If one value is missing, the prediction will be regarded as incorrect. Besides, we also find that the overall performance of \textit{Multiple Chart Understanding} is lower than \textit{Contextual Chart Understanding}. This phenomenon may be attributed to the lack of training data with multiple images as input.
\begin{table}[t]
\setlength\tabcolsep{4.3pt}
\centering
\small
\begin{tabular}{lccc}
\toprule[1pt]
\textbf{Model} & ChartQA  & DocVQA & TextVQA\\
\midrule
Donut & 41.8 & 67.5 & 43.5\\
Pix2Struct & 56.0 & 72.1 & -\\
\midrule
MiniGPT-v2 &  49.5 & 61.3& 50.7  \\
LLaVA1.5 &   52.5& 66.5& 58.2\\
Mplug-Owl & 51.6&62.2&54.3\\
\midrule
\textbf{MMCA (Ours)} &  \textbf{57.4} & \textbf{72.5}&\textbf{59.6} \\
\bottomrule[1.5pt]
\end{tabular}
\caption{Comparison with OCR-free methods and LMMs on existing public benchmarks.}
\label{tab:comparison_mmc_public}
\end{table}

\vspace{-0.1in}
\subsubsection{Results on Public Benchmarks}
\vspace{-0.05in}
We compare our \textit{MMCA} with the state-of-the-art methods on existing public benchmarks including ChartQA~\citep{masry2022chartqa}, DocVQA~\citep{mathew2021docvqa}, and TextVQA~\cite{yang2021tap}. As shown in Tab.~\ref{tab:comparison_mmc_public}, our \textit{MMCA} outperforms existing LMMs, including MiniGPT4 and LRV-Instruction, on the three public benchmarks. Moreover, without including any finetuning on the corresponding datasets, our proposed \textit{MMCA} model outperforms both Pix2Struct and Donut that are finetuned on these datasets. Such results validate that with the help of LMMs, chart understanding performance can be significantly improved.

\begin{figure*}[h]
    \centering
      \includegraphics[width=0.9\textwidth]{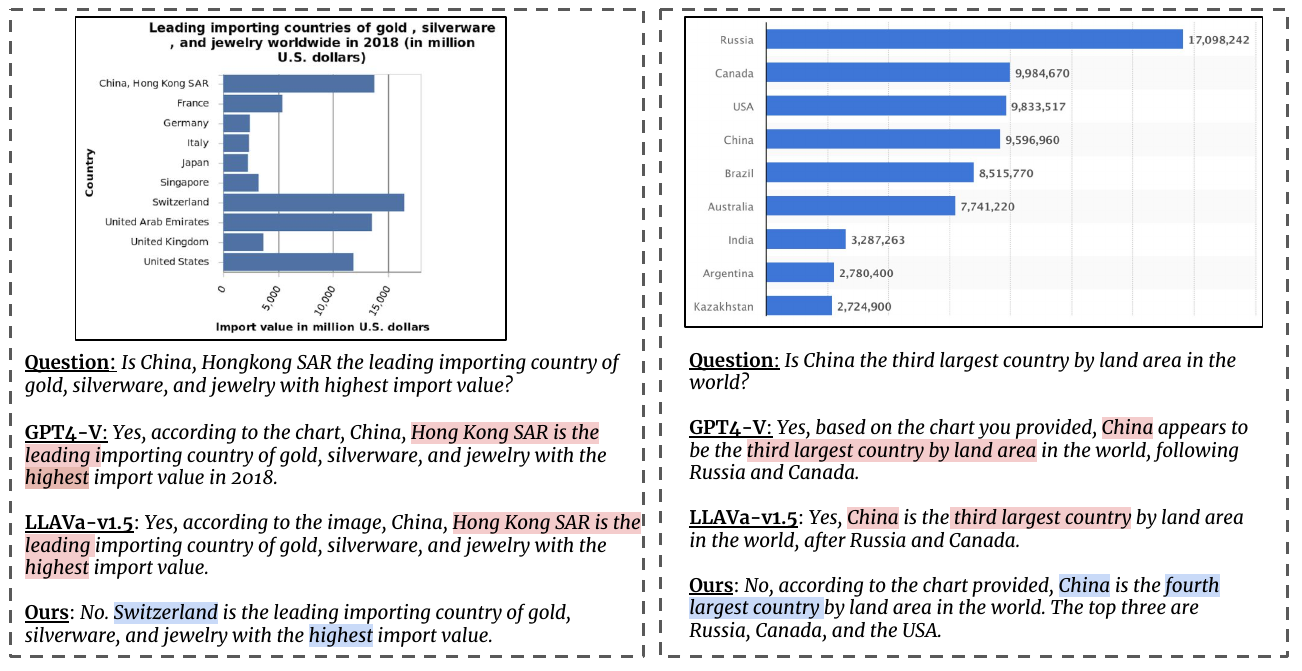}
    \caption{\textcolor{red}{RED} highlights incorrect answers while \textcolor{blue}{BLUE} highlights the correct ones. (Left): the failure of GPT-4V and LLaVA1.5 can be attributed to \textcolor{red}{\textbf{\textit{vision perception error}}}. (Right): the failure can be attributed to \textcolor{red}{\textbf{\textit{language bias}}}.}
    \label{fig:compare_gpt4}
    \vspace{-0.2in}
\end{figure*}
\vspace{-0.1in}

\subsubsection{Comparison with GPT-4V(ision)}
\vspace{-0.05in}
We further evaluate and benchmark GPT-4V~\cite{2023GPT4VisionSC} via the ``gpt-4-vision-preview'' model on our proposed \textit{MMC-Benchmark}. The quantitative results are shown in Tab.~\ref{tab:comparison_mmc} and Tab.~\ref{tab:comparison_mmc_mvqa}. Remarkably, GPT-4V surpasses all open-source LMMs by substantial margins on seven out of the nine tasks of \textit{MMC-Benchmark}. Such results prove GPT-4V's strong skills in text recognition, scientific knowledge, and math reasoning. \textbf{\textit{The only two tasks GPT-4V performs poorly are \textit{Chart to Datatable} and \textit{Chart to Json}}}. As shown in Fig.\ref{fig:gpt4_table}, GPT-4V misrecognizes the data value from the charts. GPT-4V also predicts incorrectly on the \textit{Multiple Charts Understanding} tasks such as Fig.~\ref{fig:gpt4_multiple1},~\ref{fig:gpt4_multiple2}.

We examine 100 randomly sampled error instances from GPT-4V’s predictions. The instances are analyzed by expert annotators who identify the root causes. The distribution of errors is in Fig.~\ref{fig:error_gpt4}. \textbf{Language Bias (35\%)}: As indicated in Fig.~\ref{fig:compare_gpt4} (right), the strong language prior or parametric memory misleads GPT-4V to answer \textit{``China appears to be the third largest country by land area in the world''}, which conflicts with the information mentioned in the chart \textit{``USA appears to be the third largest country by land area''}. \textbf{Perception Error (39\%)}: As in Fig.~\ref{fig:compare_gpt4} (left) and Fig.~\ref{fig:gpt4_multiple1}, the perception error occurs when GPT-4V fails to interpret the chart~\cite{liu2023hallusionbench}. The remaining errors include \textbf{Reasoning Error (15\%)} in Fig.~\ref{fig:gpt4_multiple2} and \textbf{Lack of Knowledge (11\%)} in Fig.~\ref{fig:gpt4_lack}. These errors are attributed to various factors such as complex text interpretation, lack of domain-specific knowledge, or failure to extract answers from long context. More cases are shown in Fig.~\ref{fig:gpt4_table}, and \ref{fig:gpt4_mix}.

\subsubsection{Error Analysis of Open-Source Models}
\vspace{-0.05in}
\textbf{\textit{Not Following Instructions}}. Even with a very concise instruction design, there are LMMs that do not follow the user's instructions. For example, in Fig.~\ref{fig:demo12}, when asked \textit{``Please identify the proportion of Americans who favor the coal mining.''}, PixsStruct and MiniGPT-v2 answer \textit{``Yes''} and \textit{``Most Americans favor exporting or expanding solar and wind powers.''}, respectively. In our opinion, a good chart understanding model should be able to follow instructions. However, to the best of our knowledge, most of the existing LLM-based or LMM-based models, except for GPT-4V, are not able to follow human instructions well. More examples are shown in Fig.~\ref{fig:demo11}, \ref{fig:demo13}, and \ref{fig:demo14}.

\textbf{\textit{Vision Encoder is Weak}}. Existing LMMs typically use CLIP as the vision encoder and do not update its parameters during training. However, as CLIP is trained to align visual embeddings with short captions, its capability of modeling the spatial interactions of chart elements like trend lines and color-coded legends is limited. The potential method is to add segmentation~\cite{kirillov2023segment} and project the segments into the LLM token embedding space. Instead, in our proposed \textit{MMCA} approach, we finetune LMMs on our \textit{MMC-Instruction} data by updating the vision parts during training and improving the integration of visual elements into the LLM input domain. The result improvements prove the effectiveness of \textit{MMC-Instruction} and the training strategy in \textit{MMCA}. Fig.~\ref{fig:error_open} shows the distributions of failure causes.

\label{sec:experiment}
\section{Conclusion}

This paper aims to tackle the challenge of chart understanding with Large Multimodal Models (LMMs). Firstly, we present a large-scale chart instruction-tuning dataset \textit{MMC-Instruction}, including diverse topics, language styles, chart types, and open-ended answers in line with human cognition. Secondly, we introduce a human-annotated benchmark called \textit{MMC-Benchmark} to evaluate LLMs' abilities for chart understanding quantitatively. Finally, we propose an instruction-tuned LMM called \textit{MMCA} that outperforms existing open-source SoTA methods.

\label{sec:conslusion}

\section{Limitations}
\label{sec:Limitations}
Our study innovatively utilizes a large multimodal model with 7 billion parameters, showcasing substantial capabilities within the constraints of our current computational resources. While we recognize that employing even larger models, such as the 13 billion parameter variants, could further enhance our findings, lacking access to high-end computing resources like A100 limits our current scope. This presents an exciting avenue for future research, where we aim to expand our model's complexity and depth as more advanced computational means become available.

\section{Ethical Considerations}

\textbf{Copyright and Licensing}: Strict adherence to copyright and licensing regulations is mandatory. Data from sources that prohibit copying or redistribution will be explicitly avoided. \textbf{Data Privacy}: Compliance with privacy laws and ethical standards in data handling is paramount. The annotators should avoid collecting questions that contain any private information.

\bibliography{anthology,custom}

\begin{thebibliography}{48}
\expandafter\ifx\csname natexlab\endcsname\relax\def\natexlab#1{#1}\fi

\bibitem[{Brown et~al.(2020)Brown, Mann, Ryder, Subbiah, Kaplan, Dhariwal, Neelakantan, Shyam, Sastry, Askell et~al.}]{brown2020language}
Tom Brown, Benjamin Mann, Nick Ryder, Melanie Subbiah, Jared~D Kaplan, Prafulla Dhariwal, Arvind Neelakantan, Pranav Shyam, Girish Sastry, Amanda Askell, et~al. 2020.
\newblock Language models are few-shot learners.
\newblock \emph{Advances in neural information processing systems}, 33:1877--1901.

\bibitem[{Cao et~al.(2023)Cao, Xu, Sun, Huang, and Shen}]{cao2023towards}
Yunkang Cao, Xiaohao Xu, Chen Sun, Xiaonan Huang, and Weiming Shen. 2023.
\newblock Towards generic anomaly detection and understanding: Large-scale visual-linguistic model (gpt-4v) takes the lead.
\newblock \emph{arXiv preprint arXiv:2311.02782}.

\bibitem[{Chaudhry et~al.(2020)Chaudhry, Shekhar, Gupta, Maneriker, Bansal, and Joshi}]{chaudhry2020leaf}
Ritwick Chaudhry, Sumit Shekhar, Utkarsh Gupta, Pranav Maneriker, Prann Bansal, and Ajay Joshi. 2020.
\newblock Leaf-qa: Locate, encode \& attend for figure question answering.
\newblock In \emph{Proceedings of the IEEE/CVF Winter Conference on Applications of Computer Vision}, pages 3512--3521.

\bibitem[{Chen et~al.(2023{\natexlab{a}})Chen, Zhu, Shen, Li, Liu, Zhang, Krishnamoorthi, Chandra, Xiong, and Elhoseiny}]{chen2023minigpt}
Jun Chen, Deyao Zhu, Xiaoqian Shen, Xiang Li, Zechun Liu, Pengchuan Zhang, Raghuraman Krishnamoorthi, Vikas Chandra, Yunyang Xiong, and Mohamed Elhoseiny. 2023{\natexlab{a}}.
\newblock Minigpt-v2: large language model as a unified interface for vision-language multi-task learning.
\newblock \emph{arXiv preprint arXiv:2310.09478}.

\bibitem[{Chen et~al.(2023{\natexlab{b}})Chen, Zhang, Zeng, Zhang, Zhu, and Zhao}]{chen2023shikra}
Keqin Chen, Zhao Zhang, Weili Zeng, Richong Zhang, Feng Zhu, and Rui Zhao. 2023{\natexlab{b}}.
\newblock Shikra: Unleashing multimodal llm's referential dialogue magic.
\newblock \emph{arXiv preprint arXiv:2306.15195}.

\bibitem[{Chowdhery et~al.(2022)Chowdhery, Narang, Devlin, Bosma, Mishra, Roberts, Barham, Chung, Sutton, Gehrmann et~al.}]{chowdhery2022palm}
Aakanksha Chowdhery, Sharan Narang, Jacob Devlin, Maarten Bosma, Gaurav Mishra, Adam Roberts, Paul Barham, Hyung~Won Chung, Charles Sutton, Sebastian Gehrmann, et~al. 2022.
\newblock Palm: Scaling language modeling with pathways.
\newblock \emph{arXiv preprint arXiv:2204.02311}.

\bibitem[{Dai et~al.(2023)Dai, Li, Li, Tiong, Zhao, Wang, Li, Fung, and Hoi}]{dai2023instructblip}
Wenliang Dai, Junnan Li, Dongxu Li, Anthony Meng~Huat Tiong, Junqi Zhao, Weisheng Wang, Boyang Li, Pascale Fung, and Steven Hoi. 2023.
\newblock Instructblip: Towards general-purpose vision-language models with instruction tuning.
\newblock \emph{arXiv preprint arXiv:2305.06500}.

\bibitem[{Gong et~al.(2023)Gong, Lyu, Zhang, Wang, Zheng, Zhao, Liu, Zhang, Luo, and Chen}]{gong2023multimodal}
Tao Gong, Chengqi Lyu, Shilong Zhang, Yudong Wang, Miao Zheng, Qian Zhao, Kuikun Liu, Wenwei Zhang, Ping Luo, and Kai Chen. 2023.
\newblock Multimodal-gpt: A vision and language model for dialogue with humans.
\newblock \emph{arXiv preprint arXiv:2305.04790}.

\bibitem[{Hsu et~al.(2021)Hsu, Giles, and Huang}]{hsu2021scicap}
Ting-Yao Hsu, C~Lee Giles, and Ting-Hao'Kenneth' Huang. 2021.
\newblock Scicap: Generating captions for scientific figures.
\newblock \emph{arXiv preprint arXiv:2110.11624}.

\bibitem[{Kafle et~al.(2018)Kafle, Price, Cohen, and Kanan}]{kafle2018dvqa}
Kushal Kafle, Brian Price, Scott Cohen, and Christopher Kanan. 2018.
\newblock Dvqa: Understanding data visualizations via question answering.
\newblock In \emph{Proceedings of the IEEE conference on computer vision and pattern recognition}, pages 5648--5656.

\bibitem[{Kahou et~al.(2017)Kahou, Michalski, Atkinson, K{\'a}d{\'a}r, Trischler, and Bengio}]{kahou2017figureqa}
Samira~Ebrahimi Kahou, Vincent Michalski, Adam Atkinson, {\'A}kos K{\'a}d{\'a}r, Adam Trischler, and Yoshua Bengio. 2017.
\newblock Figureqa: An annotated figure dataset for visual reasoning.
\newblock \emph{arXiv preprint arXiv:1710.07300}.

\bibitem[{Kantharaj et~al.(2022)Kantharaj, Leong, Lin, Masry, Thakkar, Hoque, and Joty}]{kantharaj2022chart}
Shankar Kantharaj, Rixie Tiffany~Ko Leong, Xiang Lin, Ahmed Masry, Megh Thakkar, Enamul Hoque, and Shafiq Joty. 2022.
\newblock Chart-to-text: A large-scale benchmark for chart summarization.
\newblock \emph{arXiv preprint arXiv:2203.06486}.

\bibitem[{Kim et~al.(2022)Kim, Hong, Yim, Nam, Park, Yim, Hwang, Yun, Han, and Park}]{kim2022ocr}
Geewook Kim, Teakgyu Hong, Moonbin Yim, JeongYeon Nam, Jinyoung Park, Jinyeong Yim, Wonseok Hwang, Sangdoo Yun, Dongyoon Han, and Seunghyun Park. 2022.
\newblock Ocr-free document understanding transformer.
\newblock In \emph{European Conference on Computer Vision}, pages 498--517. Springer.

\bibitem[{Kirillov et~al.(2023)Kirillov, Mintun, Ravi, Mao, Rolland, Gustafson, Xiao, Whitehead, Berg, Lo et~al.}]{kirillov2023segment}
Alexander Kirillov, Eric Mintun, Nikhila Ravi, Hanzi Mao, Chloe Rolland, Laura Gustafson, Tete Xiao, Spencer Whitehead, Alexander~C Berg, Wan-Yen Lo, et~al. 2023.
\newblock Segment anything.
\newblock \emph{arXiv preprint arXiv:2304.02643}.

\bibitem[{Lal et~al.(2023)Lal, Mitkari, Bhosale, and Doermann}]{lal2023lineformer}
Jay Lal, Aditya Mitkari, Mahesh Bhosale, and David Doermann. 2023.
\newblock Lineformer: Line chart data extraction using instance segmentation.
\newblock In \emph{International Conference on Document Analysis and Recognition}, pages 387--400. Springer.

\bibitem[{Lee et~al.(2023)Lee, Joshi, Turc, Hu, Liu, Eisenschlos, Khandelwal, Shaw, Chang, and Toutanova}]{lee2023pix2struct}
Kenton Lee, Mandar Joshi, Iulia~Raluca Turc, Hexiang Hu, Fangyu Liu, Julian~Martin Eisenschlos, Urvashi Khandelwal, Peter Shaw, Ming-Wei Chang, and Kristina Toutanova. 2023.
\newblock Pix2struct: Screenshot parsing as pretraining for visual language understanding.
\newblock In \emph{International Conference on Machine Learning}, pages 18893--18912. PMLR.

\bibitem[{Li et~al.(2023)Li, Li, Savarese, and Hoi}]{li2023blip}
Junnan Li, Dongxu Li, Silvio Savarese, and Steven Hoi. 2023.
\newblock Blip-2: Bootstrapping language-image pre-training with frozen image encoders and large language models.
\newblock \emph{arXiv preprint arXiv:2301.12597}.

\bibitem[{Li et~al.(2021)Li, Fu, Zhang, and Qiao}]{li2021character}
Ming Li, Bin Fu, Zhengfu Zhang, and Yu~Qiao. 2021.
\newblock Character-aware sampling and rectification for scene text recognition.
\newblock \emph{IEEE Transactions on Multimedia}, 25:649--661.

\bibitem[{Li and Tajbakhsh(2023)}]{li2023scigraphqa}
Shengzhi Li and Nima Tajbakhsh. 2023.
\newblock Scigraphqa: A large-scale synthetic multi-turn question-answering dataset for scientific graphs.
\newblock \emph{arXiv preprint arXiv:2308.03349}.

\bibitem[{Liu et~al.(2023{\natexlab{a}})Liu, Guan, Li, Chen, Yacoob, Manocha, and Zhou}]{liu2023hallusionbench}
Fuxiao Liu, Tianrui Guan, Zongxia Li, Lichang Chen, Yaser Yacoob, Dinesh Manocha, and Tianyi Zhou. 2023{\natexlab{a}}.
\newblock Hallusionbench: You see what you think? or you think what you see? an image-context reasoning benchmark challenging for gpt-4v (ision), llava-1.5, and other multi-modality models.
\newblock \emph{arXiv preprint arXiv:2310.14566}.

\bibitem[{Liu et~al.(2023{\natexlab{b}})Liu, Lin, Li, Wang, Yacoob, and Wang}]{liu2023aligning}
Fuxiao Liu, Kevin Lin, Linjie Li, Jianfeng Wang, Yaser Yacoob, and Lijuan Wang. 2023{\natexlab{b}}.
\newblock Aligning large multi-modal model with robust instruction tuning.
\newblock \emph{arXiv preprint arXiv:2306.14565}.

\bibitem[{Liu et~al.(2023{\natexlab{c}})Liu, Tan, and Tensmeyer}]{liu2023documentclip}
Fuxiao Liu, Hao Tan, and Chris Tensmeyer. 2023{\natexlab{c}}.
\newblock Documentclip: Linking figures and main body text in reflowed documents.
\newblock \emph{arXiv preprint arXiv:2306.06306}.

\bibitem[{Liu et~al.(2020)Liu, Wang, Wang, and Ordonez}]{liu2020visualnews}
Fuxiao Liu, Yinghan Wang, Tianlu Wang, and Vicente Ordonez. 2020.
\newblock \href {http://arxiv.org/abs/2010.03743} {Visualnews : Benchmark and challenges in entity-aware image captioning}.

\bibitem[{Liu et~al.(2023{\natexlab{d}})Liu, Li, Li, and Lee}]{liu2023improved}
Haotian Liu, Chunyuan Li, Yuheng Li, and Yong~Jae Lee. 2023{\natexlab{d}}.
\newblock Improved baselines with visual instruction tuning.
\newblock \emph{arXiv preprint arXiv:2310.03744}.

\bibitem[{Liu et~al.(2023{\natexlab{e}})Liu, Li, Wu, and Lee}]{liu2023visual}
Haotian Liu, Chunyuan Li, Qingyang Wu, and Yong~Jae Lee. 2023{\natexlab{e}}.
\newblock Visual instruction tuning.
\newblock \emph{arXiv preprint arXiv:2304.08485}.

\bibitem[{Manyika(2023)}]{manyika2023overview}
James Manyika. 2023.
\newblock An overview of bard: an early experiment with generative ai.
\newblock \emph{AI. Google Static Documents}.

\bibitem[{Masry et~al.(2023)Masry, Kavehzadeh, Do, Hoque, and Joty}]{masry2023unichart}
Ahmed Masry, Parsa Kavehzadeh, Xuan~Long Do, Enamul Hoque, and Shafiq Joty. 2023.
\newblock Unichart: A universal vision-language pretrained model for chart comprehension and reasoning.
\newblock \emph{arXiv preprint arXiv:2305.14761}.

\bibitem[{Masry et~al.(2022)Masry, Long, Tan, Joty, and Hoque}]{masry2022chartqa}
Ahmed Masry, Do~Xuan Long, Jia~Qing Tan, Shafiq Joty, and Enamul Hoque. 2022.
\newblock Chartqa: A benchmark for question answering about charts with visual and logical reasoning.
\newblock \emph{arXiv preprint arXiv:2203.10244}.

\bibitem[{Mathew et~al.(2021)Mathew, Karatzas, and Jawahar}]{mathew2021docvqa}
Minesh Mathew, Dimosthenis Karatzas, and CV~Jawahar. 2021.
\newblock Docvqa: A dataset for vqa on document images.
\newblock In \emph{Proceedings of the IEEE/CVF winter conference on applications of computer vision}, pages 2200--2209.

\bibitem[{Methani et~al.(2020)Methani, Ganguly, Khapra, and Kumar}]{methani2020plotqa}
Nitesh Methani, Pritha Ganguly, Mitesh~M Khapra, and Pratyush Kumar. 2020.
\newblock Plotqa: Reasoning over scientific plots.
\newblock In \emph{Proceedings of the IEEE/CVF Winter Conference on Applications of Computer Vision}, pages 1527--1536.

\bibitem[{OpenAI(2022)}]{openai2022ChatGPT}
OpenAI. 2022.
\newblock \href {https://openai.com/blog/chatgpt} {Introducing chatgpt}.

\bibitem[{OpenAI(2023{\natexlab{a}})}]{openai2023GPT4}
OpenAI. 2023{\natexlab{a}}.
\newblock Gpt-4 technical report.
\newblock \emph{arXiv preprint arXiv:2303.08774}.

\bibitem[{OpenAI(2023{\natexlab{b}})}]{2023GPT4VisionSC}
OpenAI. 2023{\natexlab{b}}.
\newblock \href {https://cdn.openai.com/papers/GPTV_System_Card.pdf} {Gpt-4v(ision) system card}.

\bibitem[{Tang et~al.(2023)Tang, Boggust, and Satyanarayan}]{tang2023vistext}
Benny~J Tang, Angie Boggust, and Arvind Satyanarayan. 2023.
\newblock Vistext: A benchmark for semantically rich chart captioning.
\newblock \emph{arXiv preprint arXiv:2307.05356}.

\bibitem[{Touvron et~al.(2023)Touvron, Lavril, Izacard, Martinet, Lachaux, Lacroix, Rozi{\`e}re, Goyal, Hambro, Azhar et~al.}]{touvron2023llama}
Hugo Touvron, Thibaut Lavril, Gautier Izacard, Xavier Martinet, Marie-Anne Lachaux, Timoth{\'e}e Lacroix, Baptiste Rozi{\`e}re, Naman Goyal, Eric Hambro, Faisal Azhar, et~al. 2023.
\newblock Llama: Open and efficient foundation language models.
\newblock \emph{arXiv preprint arXiv:2302.13971}.

\bibitem[{Wu et~al.(2023{\natexlab{a}})Wu, Yin, Qi, Wang, Tang, and Duan}]{wu2023visual}
Chenfei Wu, Shengming Yin, Weizhen Qi, Xiaodong Wang, Zecheng Tang, and Nan Duan. 2023{\natexlab{a}}.
\newblock Visual chatgpt: Talking, drawing and editing with visual foundation models.
\newblock \emph{arXiv preprint arXiv:2303.04671}.

\bibitem[{Wu et~al.(2023{\natexlab{b}})Wu, Fei, Qu, Ji, and Chua}]{wu2023next}
Shengqiong Wu, Hao Fei, Leigang Qu, Wei Ji, and Tat-Seng Chua. 2023{\natexlab{b}}.
\newblock Next-gpt: Any-to-any multimodal llm.
\newblock \emph{arXiv preprint arXiv:2309.05519}.

\bibitem[{Xu et~al.(2024)Xu, Li, Tao, Shen, Cheng, Li, Xu, Tao, and Zhou}]{xu2024survey}
Xiaohan Xu, Ming Li, Chongyang Tao, Tao Shen, Reynold Cheng, Jinyang Li, Can Xu, Dacheng Tao, and Tianyi Zhou. 2024.
\newblock A survey on knowledge distillation of large language models.
\newblock \emph{arXiv preprint arXiv:2402.13116}.

\bibitem[{Yang et~al.(2023{\natexlab{a}})Yang, Song, Li, Zhao, Ge, Li, and Shan}]{yang2023gpt4tools}
Rui Yang, Lin Song, Yanwei Li, Sijie Zhao, Yixiao Ge, Xiu Li, and Ying Shan. 2023{\natexlab{a}}.
\newblock Gpt4tools: Teaching large language model to use tools via self-instruction.
\newblock \emph{arXiv preprint arXiv:2305.18752}.

\bibitem[{Yang et~al.(2023{\natexlab{b}})Yang, Li, Lin, Wang, Lin, Liu, and Wang}]{yang2023dawn}
Zhengyuan Yang, Linjie Li, Kevin Lin, Jianfeng Wang, Chung-Ching Lin, Zicheng Liu, and Lijuan Wang. 2023{\natexlab{b}}.
\newblock The dawn of lmms: Preliminary explorations with gpt-4v (ision).
\newblock \emph{arXiv preprint arXiv:2309.17421}, 9.

\bibitem[{Yang et~al.(2023{\natexlab{c}})Yang, Li, Wang, Lin, Azarnasab, Ahmed, Liu, Liu, Zeng, and Wang}]{yang2023mm}
Zhengyuan Yang, Linjie Li, Jianfeng Wang, Kevin Lin, Ehsan Azarnasab, Faisal Ahmed, Zicheng Liu, Ce~Liu, Michael Zeng, and Lijuan Wang. 2023{\natexlab{c}}.
\newblock Mm-react: Prompting chatgpt for multimodal reasoning and action.
\newblock \emph{arXiv preprint arXiv:2303.11381}.

\bibitem[{Yang et~al.(2021)Yang, Lu, Wang, Yin, Florencio, Wang, Zhang, Zhang, and Luo}]{yang2021tap}
Zhengyuan Yang, Yijuan Lu, Jianfeng Wang, Xi~Yin, Dinei Florencio, Lijuan Wang, Cha Zhang, Lei Zhang, and Jiebo Luo. 2021.
\newblock Tap: Text-aware pre-training for text-vqa and text-caption.
\newblock In \emph{Proceedings of the IEEE/CVF conference on computer vision and pattern recognition}, pages 8751--8761.

\bibitem[{Ye et~al.(2023)Ye, Xu, Xu, Ye, Yan, Zhou, Wang, Hu, Shi, Shi et~al.}]{ye2023mplug}
Qinghao Ye, Haiyang Xu, Guohai Xu, Jiabo Ye, Ming Yan, Yiyang Zhou, Junyang Wang, Anwen Hu, Pengcheng Shi, Yaya Shi, et~al. 2023.
\newblock mplug-owl: Modularization empowers large language models with multimodality.
\newblock \emph{arXiv preprint arXiv:2304.14178}.

\bibitem[{Yin et~al.(2023)Yin, Fu, Zhao, Li, Sun, Xu, and Chen}]{yin2023survey}
Shukang Yin, Chaoyou Fu, Sirui Zhao, Ke~Li, Xing Sun, Tong Xu, and Enhong Chen. 2023.
\newblock A survey on multimodal large language models.
\newblock \emph{arXiv preprint arXiv:2306.13549}.

\bibitem[{Yu et~al.(2023)Yu, Yang, Li, Wang, Lin, Liu, Wang, and Wang}]{yu2023mm}
Weihao Yu, Zhengyuan Yang, Linjie Li, Jianfeng Wang, Kevin Lin, Zicheng Liu, Xinchao Wang, and Lijuan Wang. 2023.
\newblock Mm-vet: Evaluating large multimodal models for integrated capabilities.
\newblock \emph{arXiv preprint arXiv:2308.02490}.

\bibitem[{Zhai et~al.(2023)Zhai, Tong, Li, Cai, Qu, Lee, and Ma}]{zhai2023investigating}
Yuexiang Zhai, Shengbang Tong, Xiao Li, Mu~Cai, Qing Qu, Yong~Jae Lee, and Yi~Ma. 2023.
\newblock Investigating the catastrophic forgetting in multimodal large language models.
\newblock \emph{arXiv preprint arXiv:2309.10313}.

\bibitem[{Zhang et~al.(2023)Zhang, Fei, Yao, Ji, Li, Liu, and Chua}]{zhang2023transfer}
Ao~Zhang, Hao Fei, Yuan Yao, Wei Ji, Li~Li, Zhiyuan Liu, and Tat-Seng Chua. 2023.
\newblock Transfer visual prompt generator across llms.
\newblock \emph{arXiv preprint arXiv:2305.01278}.

\bibitem[{Zhu et~al.(2023)Zhu, Chen, Shen, Li, and Elhoseiny}]{zhu2023minigpt}
Deyao Zhu, Jun Chen, Xiaoqian Shen, Xiang Li, and Mohamed Elhoseiny. 2023.
\newblock Minigpt-4: Enhancing vision-language understanding with advanced large language models.
\newblock \emph{arXiv preprint arXiv:2304.10592}.

\end{thebibliography}

\clearpage
\appendix
\section{Appendix}

\subsection{MMC-Benchmark}
In this section, we discuss more about our \textit{MMC-Benchmark}.

\begin{figure*}[h]
    \centering
      \includegraphics[width=0.9\textwidth]{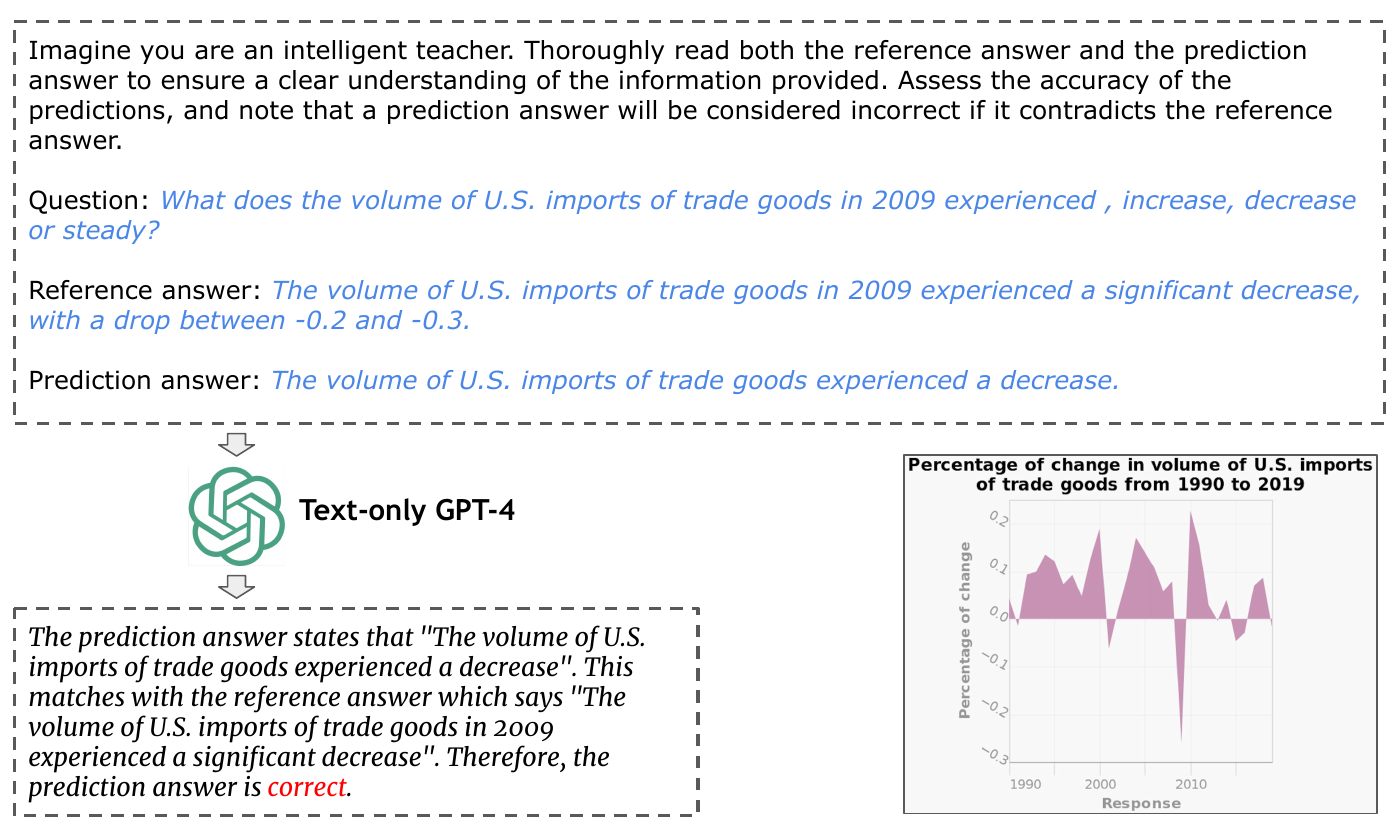}
    \caption{An example of Generative Ability Evaluation by \textit{text-only} GPT-4. In the prompt, we provide GPT-4 with the question, reference answer, and predictions from models. Then, GPT-4 accesses the accuracy of the model prediction following our instruction.}
    \label{fig:prompt_chatgpt}
\end{figure*}

\textbf{Generation Ability Evaluation} utilizes GPT-4 to assess the accuracy of the model prediction given the question and reference answers in Fig.~\ref{fig:prompt_chatgpt}. Then we ask GPT-4 to assess the prediction accuracy.

\begin{figure}[t]
    \centering
      \includegraphics[width=0.45\textwidth]{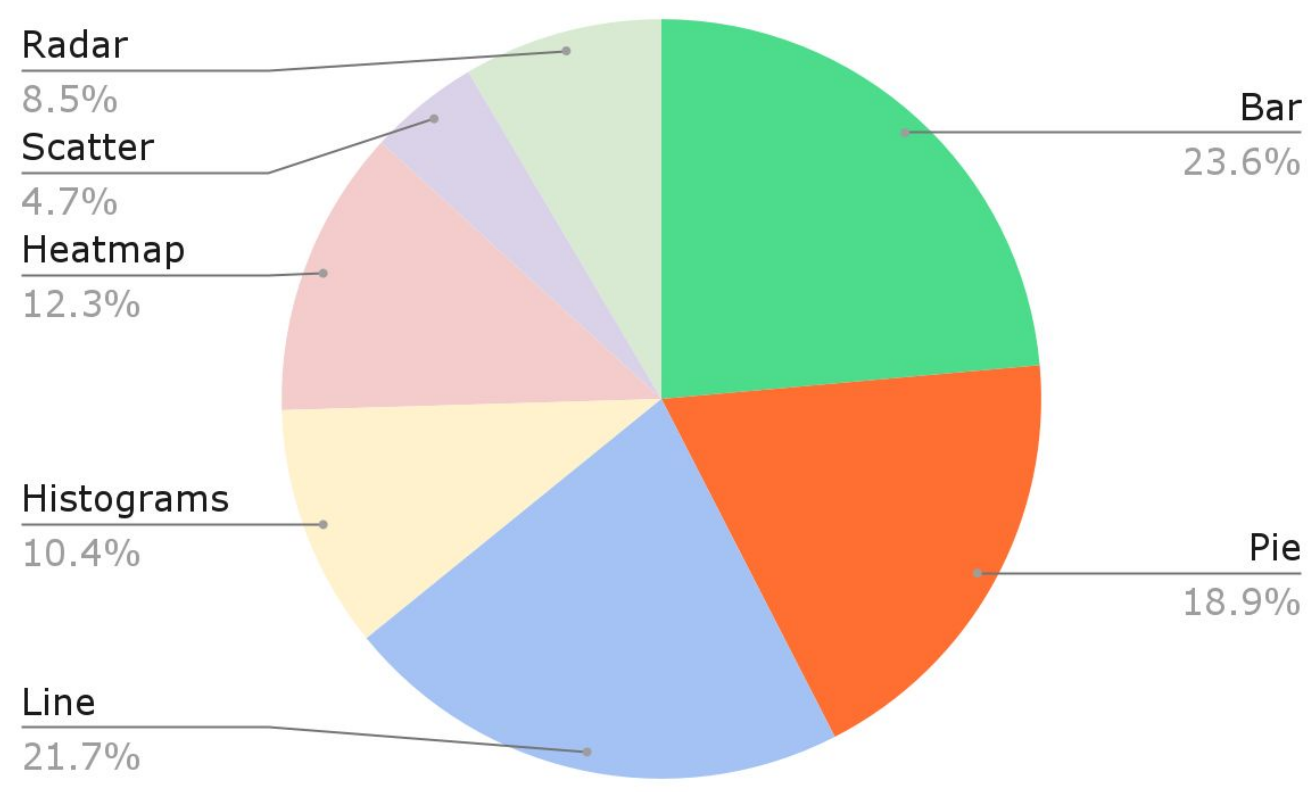}
    \caption{Distributions of chart types in \textit{MMC-Benchmark}.}
    \label{fig:type_stat}
\end{figure}

\begin{figure}[t]
    \centering
      \includegraphics[width=0.45\textwidth]{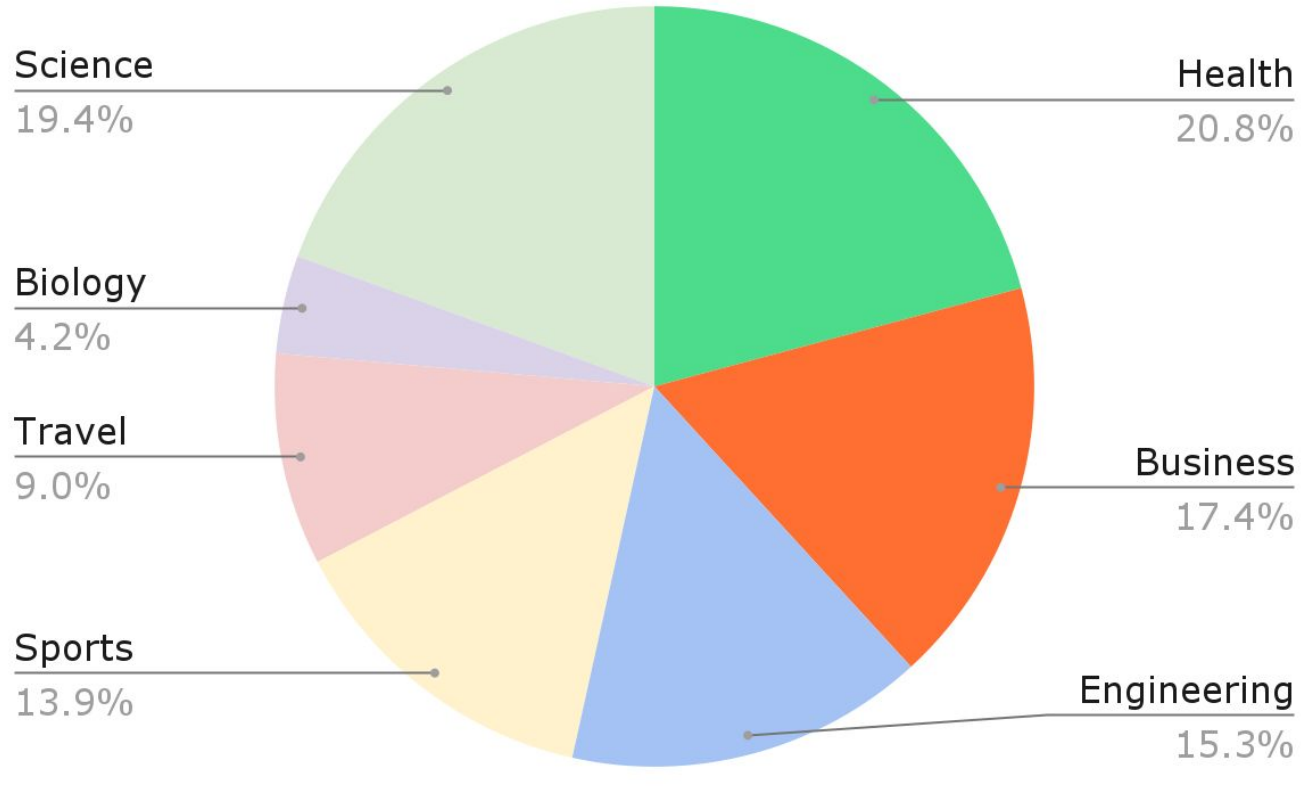}
    \caption{Distributions of chart topic in \textit{MMC-Benchmark}.}
    \label{fig:topic_stat}
\end{figure}

\begin{figure*}[h]
    \centering
      \includegraphics[width=1\textwidth]{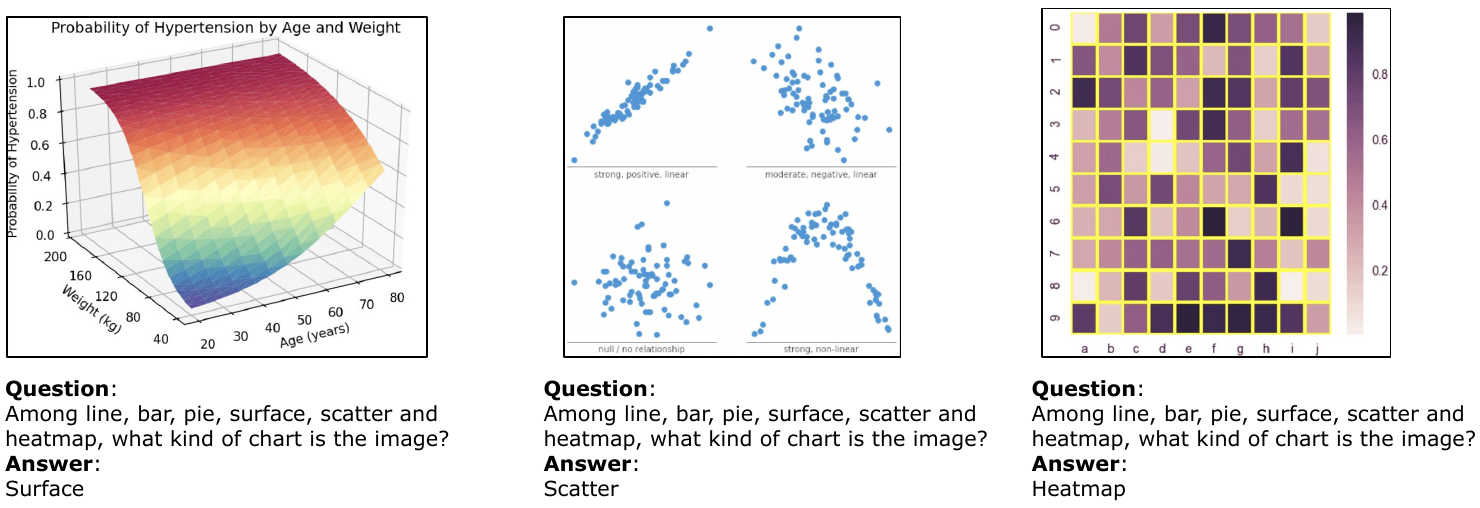}
    \caption{Examples of the \textit{Chart Type Classification} task.}
    \label{fig:type_example}
\end{figure*}

\begin{figure*}[h]
    \centering
      \includegraphics[width=1\textwidth]{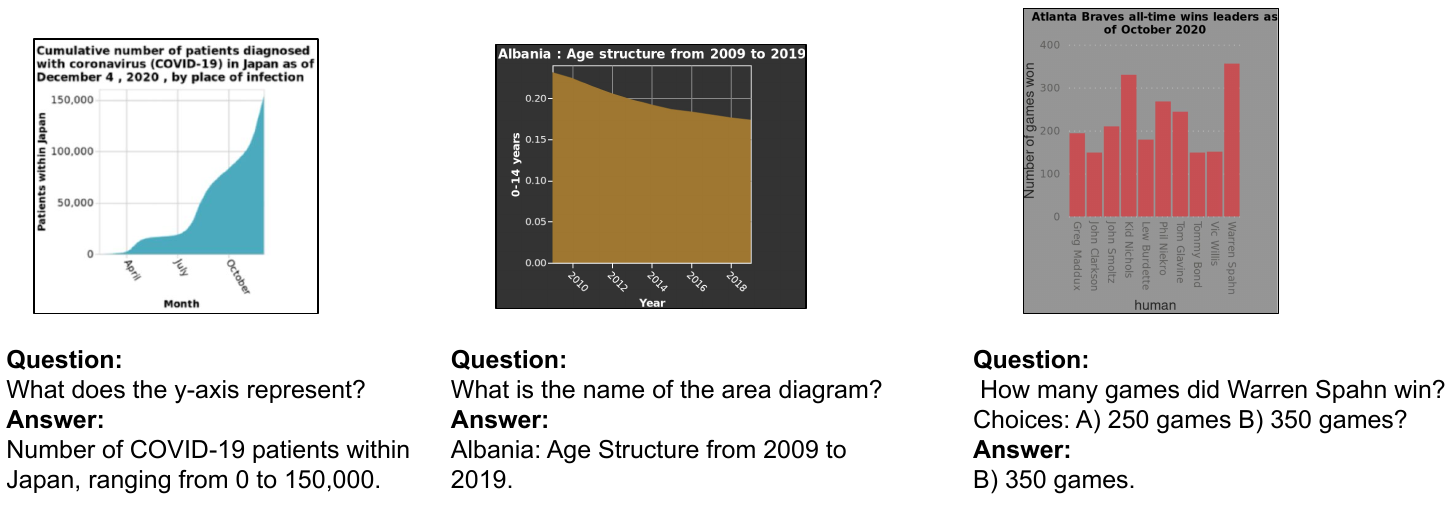}
    \caption{Examples of the \textit{Chart Information Extraction} task.}
    \label{fig:IE_example}
\end{figure*}

\begin{figure*}[h]
    \centering
      \includegraphics[width=1\textwidth]{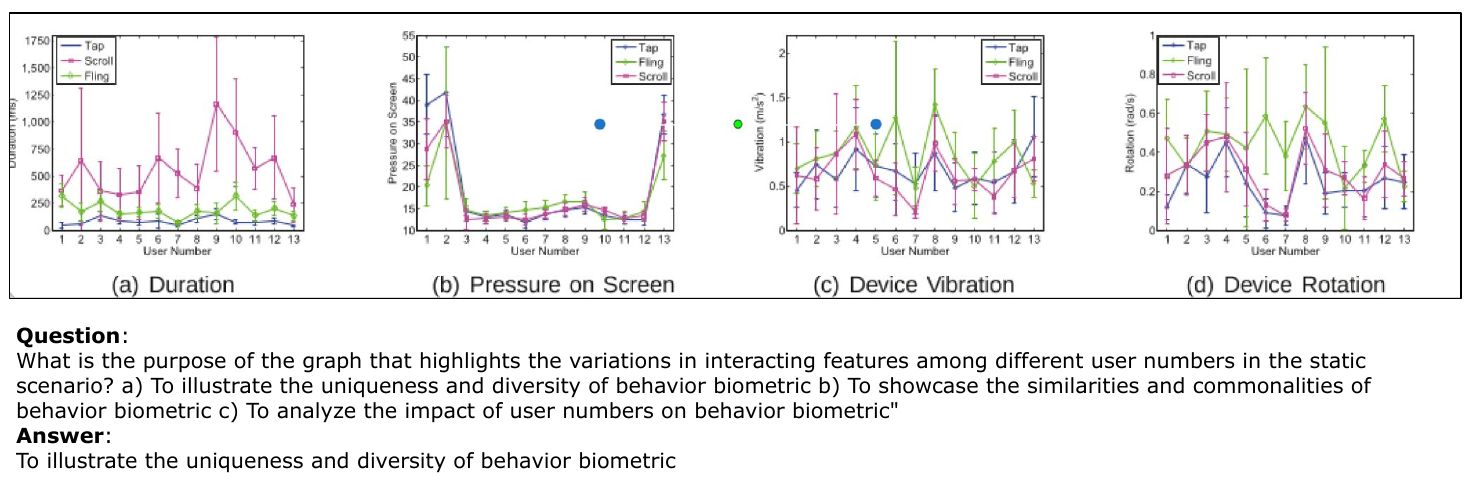}
    \caption{Examples of the \textit{Multiple Charts Understanding} task.}
    \label{fig:multiple_example}
\end{figure*}

\begin{figure*}[h]
    \centering
      \includegraphics[width=1\textwidth]{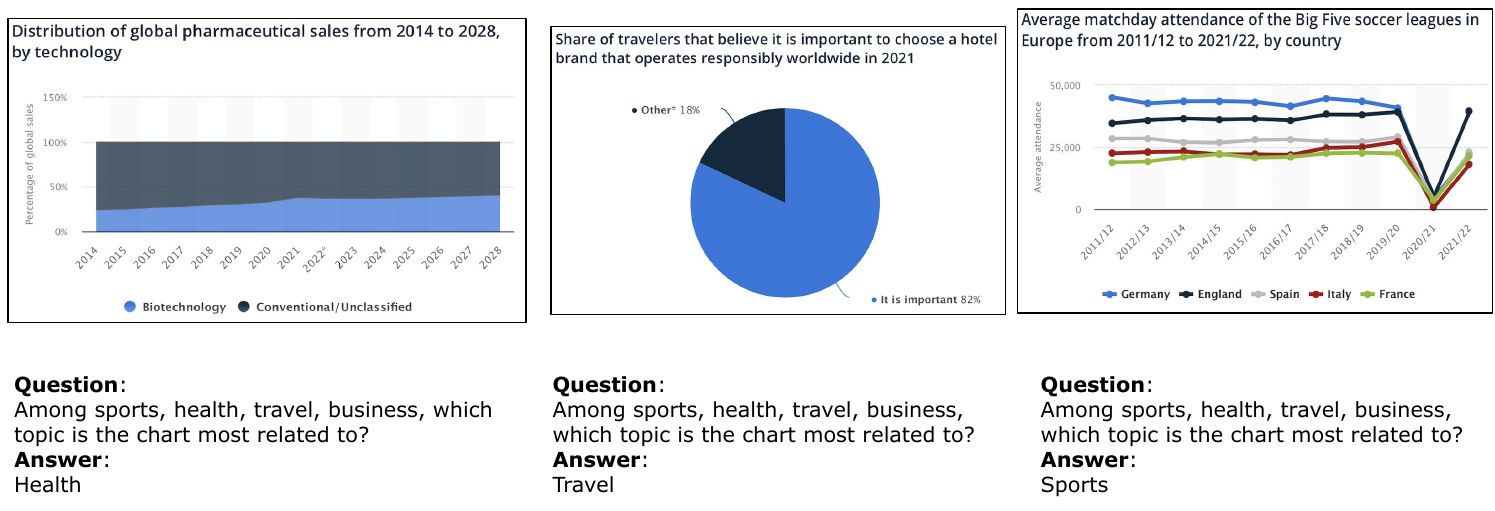}
    \caption{Examples of the \textit{Chart Topic Classification} task.}
    \label{fig:topic_example}
\end{figure*}

\begin{figure*}[h]
    \centering
      \includegraphics[width=1\textwidth]{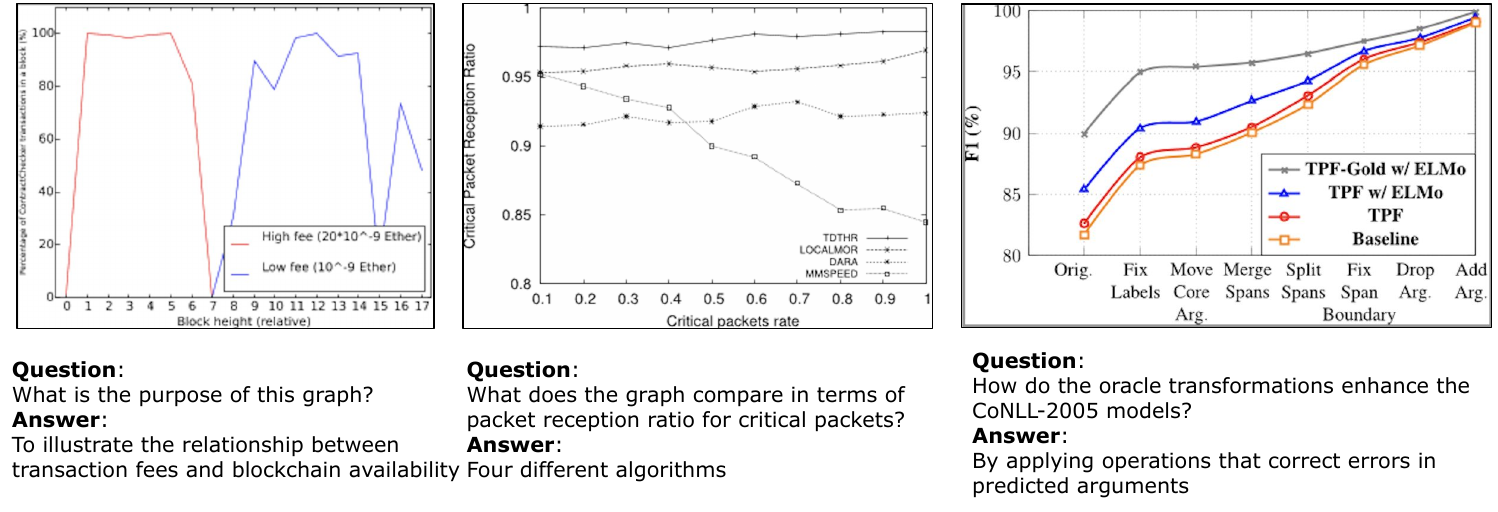}
    \caption{Examples of the \textit{Arxiv Chart Understanding} task.}
    \label{fig:arxiv_example}
\end{figure*}

\begin{figure*}[h]
    \centering
      \includegraphics[width=1\textwidth]{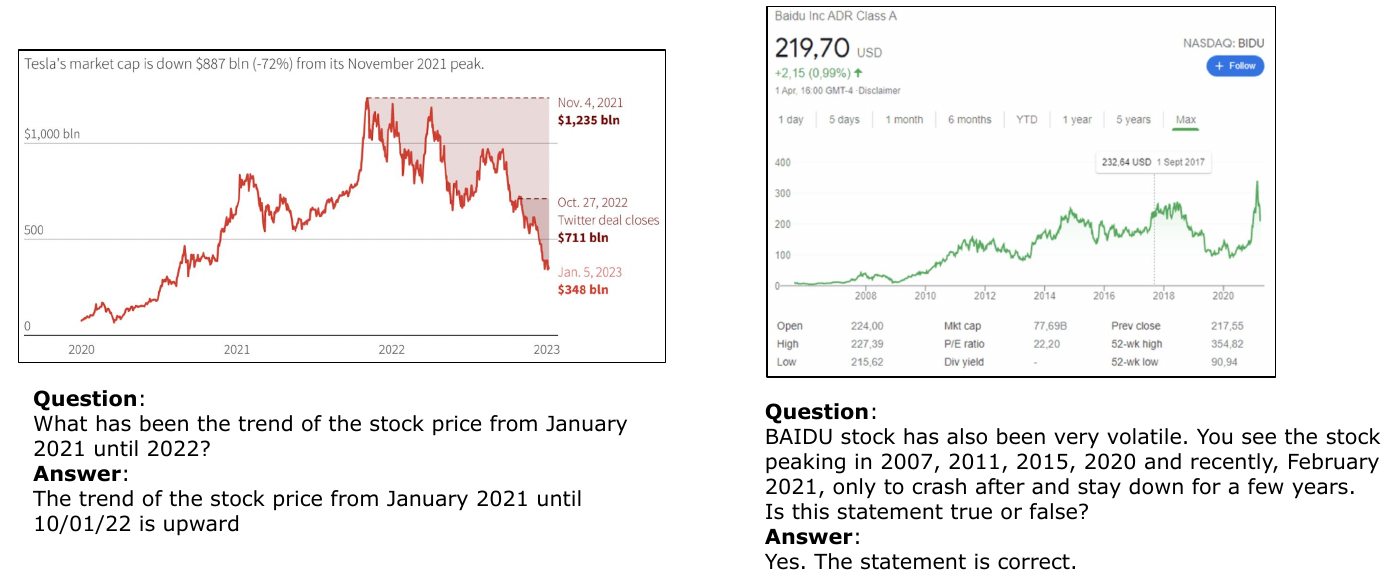}
    \caption{Examples of the \textit{Stock Charts Analysis} task.}
    \label{fig:stock_example}
\end{figure*}

\begin{figure*}[h]
    \centering
      \includegraphics[width=1\textwidth]{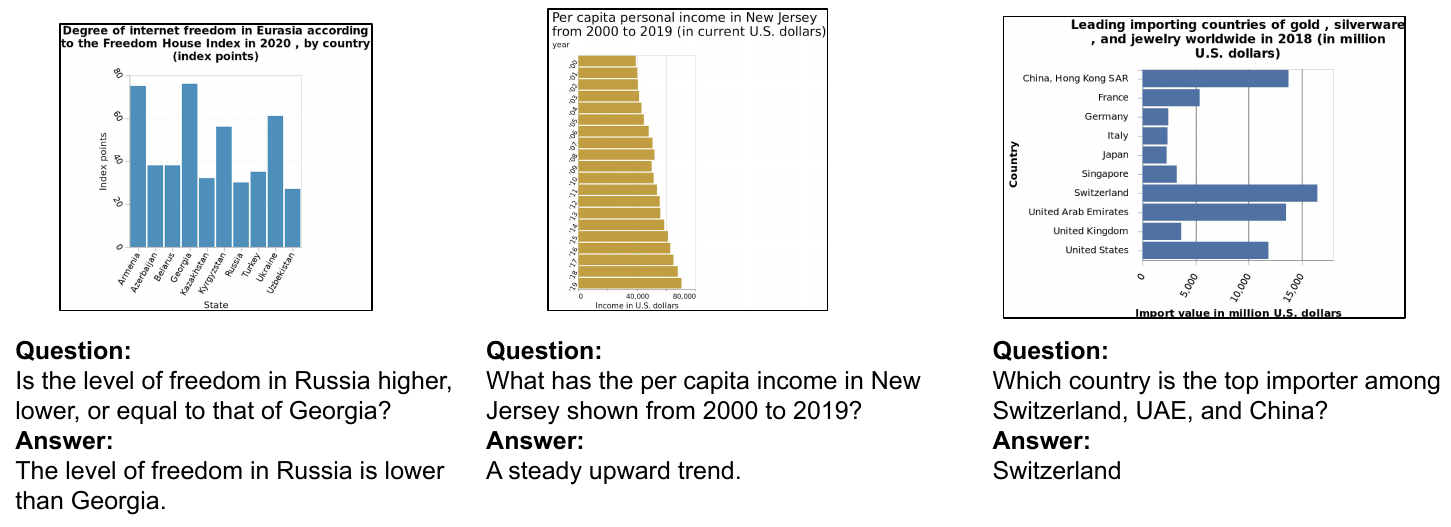}
    \caption{Examples of the \textit{Chart Reasoning} task.}
    \label{fig:CR_example}
\end{figure*}

\begin{figure*}[h]
    \centering
      \includegraphics[width=0.8\textwidth]{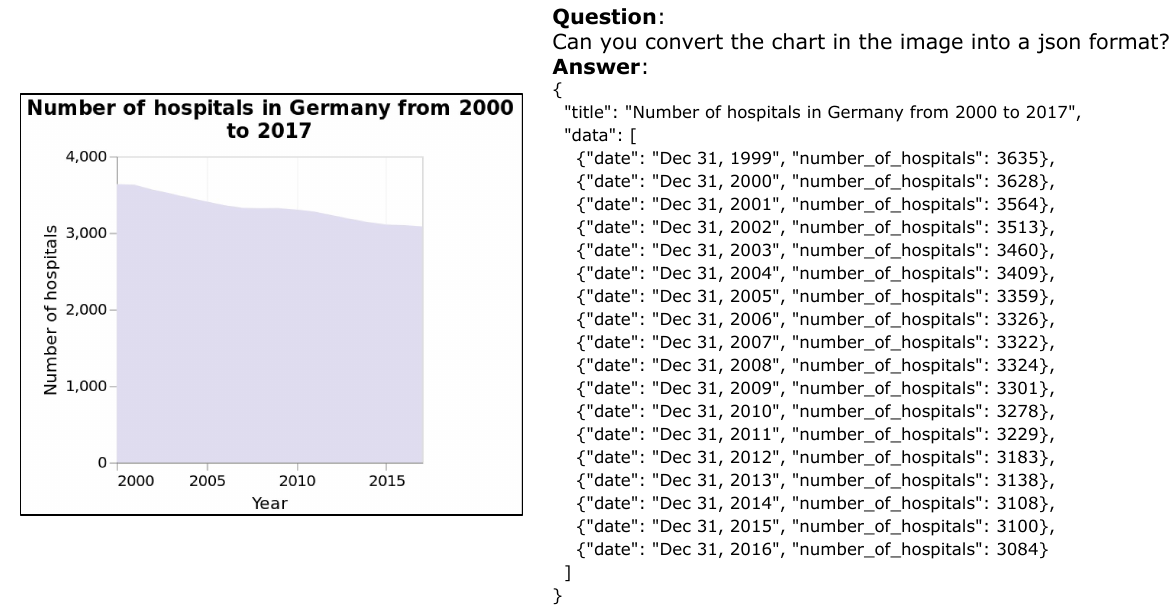}
    \caption{Examples of the \textit{Chart to Json} task.}
    \label{fig:json_example}
\end{figure*}

\textbf{Distriutions of Plot Types and Topics.} Fig.~\ref{fig:topic_stat} and Fig.~\ref{fig:type_stat} present the distributions of chart topic and plot types in \textit{MMC-Benchmark}. Fig.~\ref{fig:type_example}, Fig.~\ref{fig:IE_example}, Fig.~\ref{fig:multiple_example}, Fig.~\ref{fig:topic_example}, Fig.~\ref{fig:arxiv_example}, Fig.~\ref{fig:stock_example}, Fig.~\ref{fig:CR_example} and Fig.~\ref{fig:json_example} show the data examples of different tasks in our \textit{MMC-Benchmark.}

\subsection{Experiment}
\subsubsection{More Experiments Results}
We further compare \textit{MMCA} with Donut~\citep{kim2022ocr}, BLIP-2~\cite{li2023blip}, InstructBLIP~\cite{dai2023instructblip} and Shikra~\cite{chen2023shikra}. From Tab.~\ref{tab:comparison_mmc_mvqa_more}, we observe that non-LLM based models like Donut work well on the \textit{Chart Information Extraction} and \textit{Chart Reasoning} tasks. However, the performance drops a lot when facing other tasks, including \textit{Multiple Chart Understanding}, \textit{Chart Type Classification}, and \textit{Chart to Json}. There could be two reasons. First, the language decoder of non-LLM can not understand the questions correctly. Second, Donut's training set is not diverse enough to cover various topics and plot types. It demonstrates the value of our \textit{MMC-Instruction}.

\subsubsection{Implementation Details}
Our \textit{MMCA} model is trained with 8 Nvidia Tesla V100 GPUs. Based on the second-stage checkpoint of mPLUG-Owl, we conduct Chart Text Alignment training for one epoch with a batch size of 8. We use the same data augmentation strategy as in BLIP-2~\cite{li2023blip}, including random resized cropping and horizontal flipping with a probability of 0.5. The number of learnable queries is set to 64. We use the AdamW optimizer. The cosine learning rate decay scheduler is used with a peak learning rate of $1e^{-4}$ and 1,000 warmup steps. For the learning rate of the vision encoder, we employ layer-wise learning rate decay with a factor of 0.9 to retain the low-level visual representation. For Chart Instruction Turning, we train the language model for three epochs with a learning rate of $2e^{-5}$ and a batch size of 8.

\subsubsection{Multiple-Choice Questions Evaluation}
For multiple-choice questions, we design systematic, rule-based evaluation pipelines. Specifically, we construct robust regular expressions and develop response-processing workflows to mitigate the potential influence of any intermediate generations (e.g., reasoning steps, calculations) in the long response. These are employed to extract key phrases, such as numbers and conclusion phrases, from the long responses for accurate answer matching. If there is no valid answer in the model’s response, we perform random selection as a remedy for multiple-choice questions or consider the response incorrect for open questions.

\subsubsection{Error Analysis of GPT-4V(ision)}
\begin{figure}[t]
    \centering
      \includegraphics[width=0.45\textwidth]{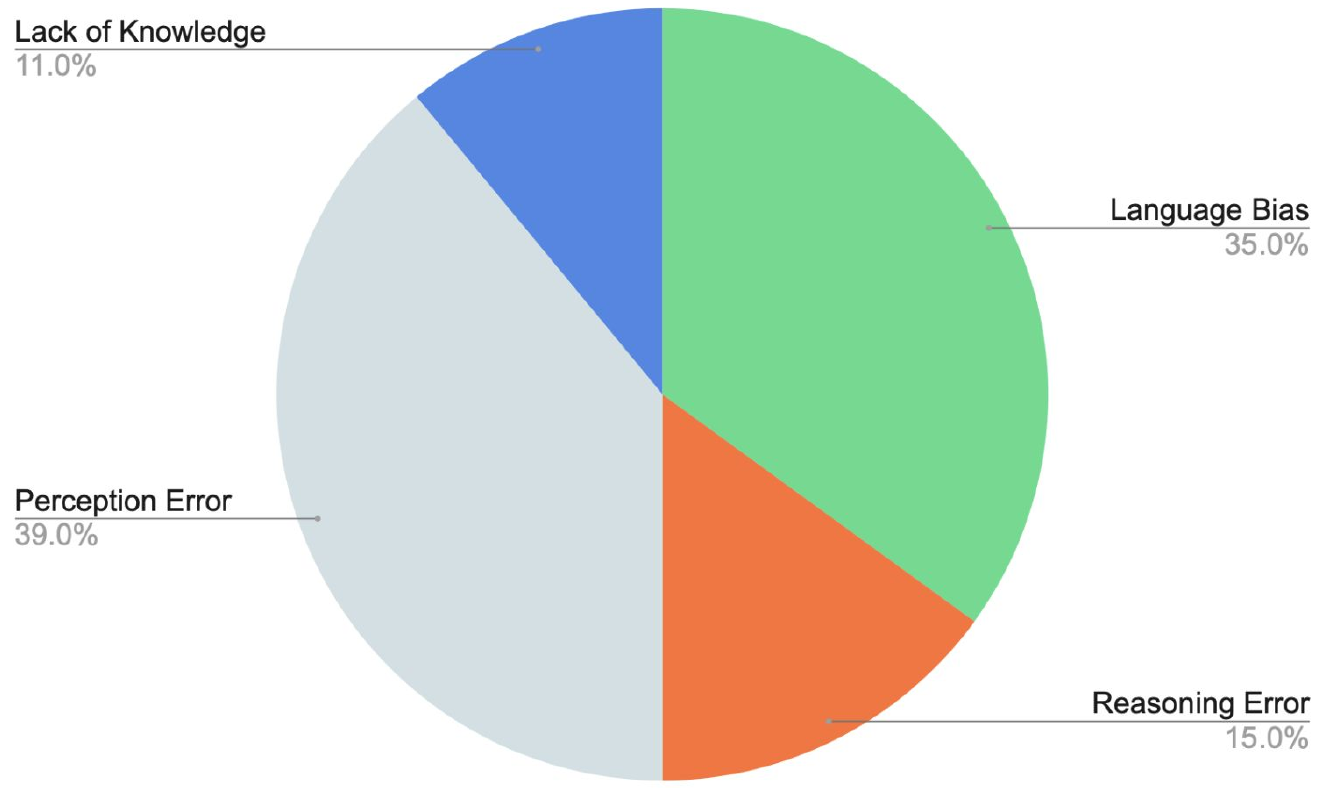}
    \caption{Error distribution of GPT-4V over 100 randomly sampled error instances.}
    \label{fig:error_gpt4}
\end{figure}

We examine 100 randomly sampled error instances from GPT-4V’s predictions. The instances are analyzed by expert annotators who identify the root causes. The distribution of errors is in Fig.~\ref{fig:error_gpt4}. 

\textbf{Language Bias (35\%).} \textit{Language Bias} refers to perceptions formed without relevant visual input. As indicated in Fig.~\ref{fig:compare_gpt4} (right), the strong language prior or parametric memory misleads GPT-4V to answer \textit{``China appears to be the third largest country by land area in the world''}, which conflicts with the information mentioned in the chart \textit{``USA appears to be the third largest country by land area''}.

\begin{figure*}[h]
    \centering
      \includegraphics[width=0.9\textwidth]{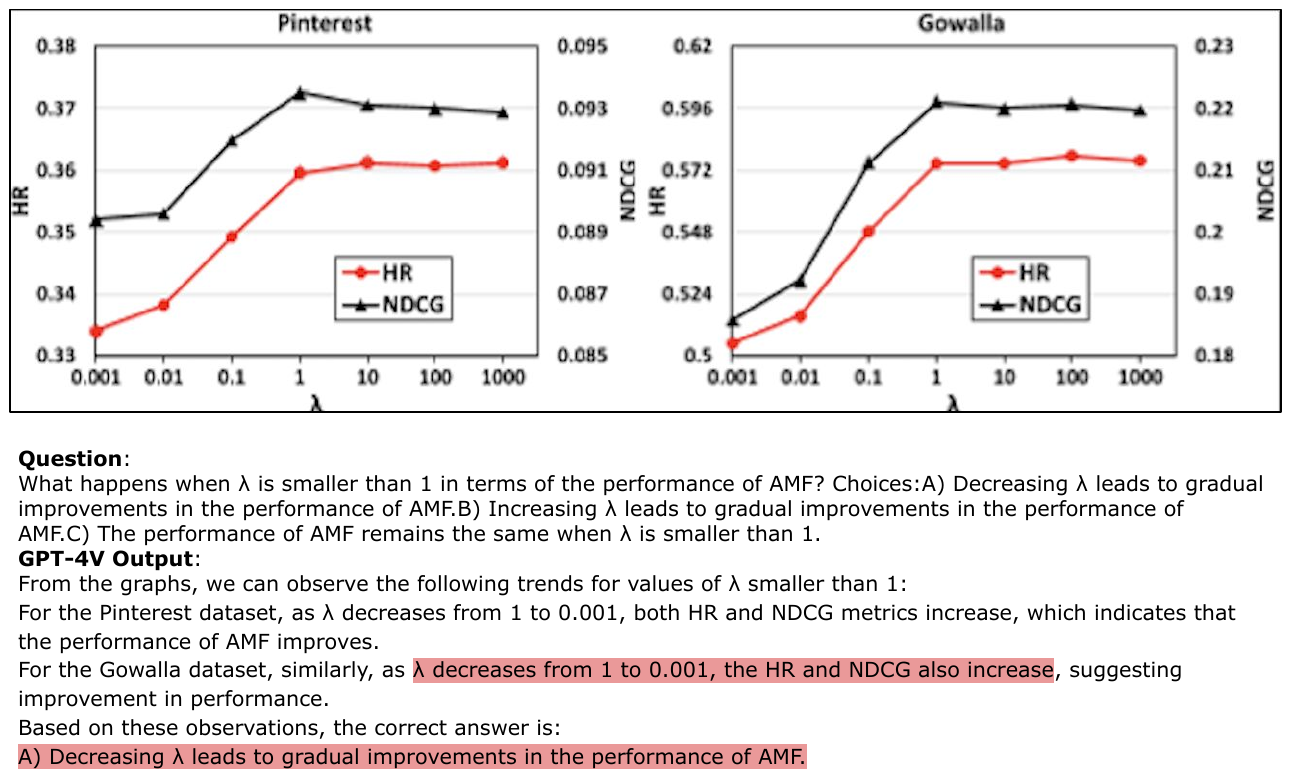}
    \caption{Failure cases of GPT-4V on \textit{Multiple Chart Understanding} task. \textcolor{red}{RED} means the incorrect answers from GPT-4V. \textbf{Error category}: \textcolor{red}{\textbf{\textit{Vision Perception Error}}}.}
    \label{fig:gpt4_multiple1}
\end{figure*}

\textbf{Perception Error (39\%).} \textit{Perception Error} denotes the misinterpretation of accurate visual information. As depicted in Fig.~\ref{fig:compare_gpt4} (left), the perception error occurs when GPT-4V fails to detect the trend in the chart (Fig.~\ref{fig:gpt4_multiple1}).

\begin{figure*}[h]
    \centering
      \includegraphics[width=0.9\textwidth]{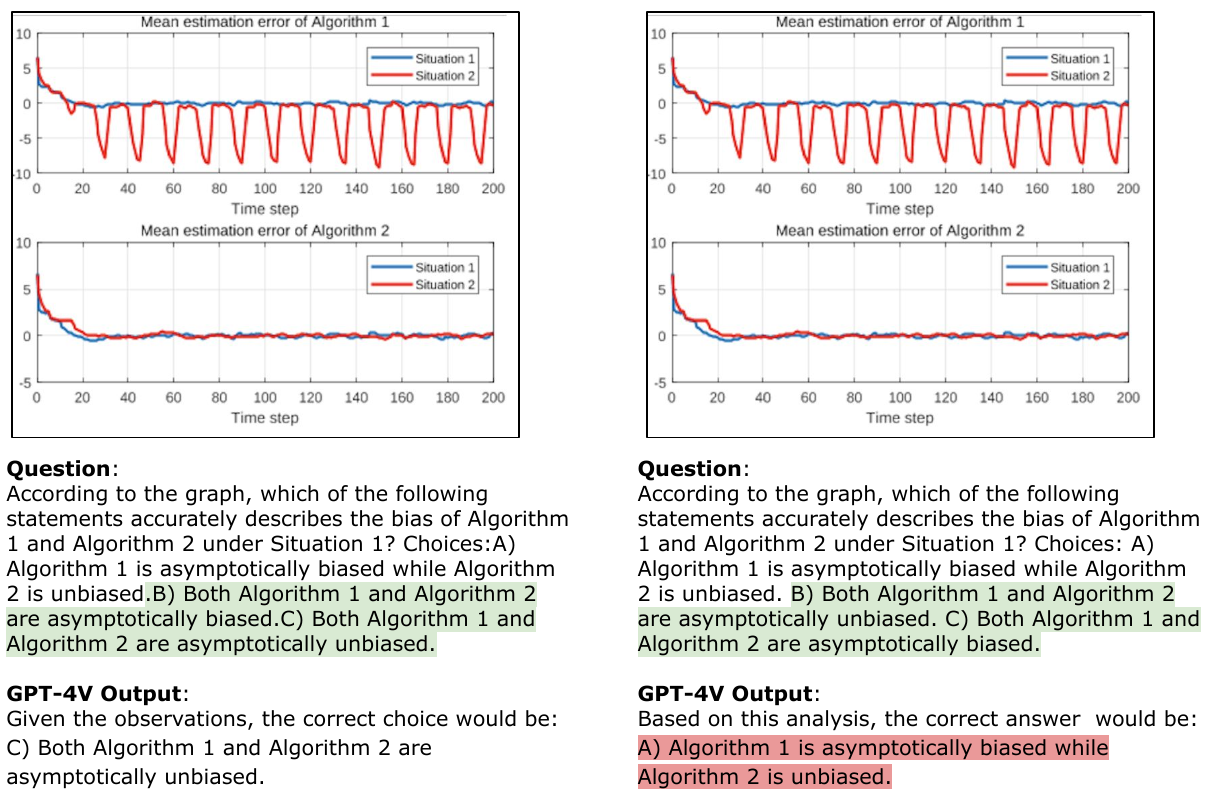}
    \caption{Failure cases of GPT-4V on \textit{Multiple Chart Understanding} task. \textcolor{red}{RED} means the incorrect answers from GPT-4V. \textcolor{green}{GREEN} denotes the section of the question where the order of choices is switched. \textbf{Error category}: \textcolor{red}{\textbf{\textit{Reasoning Error}}}.}
    \label{fig:gpt4_multiple2}
\end{figure*}

\begin{figure*}[h]
    \centering
      \includegraphics[width=0.9\textwidth]{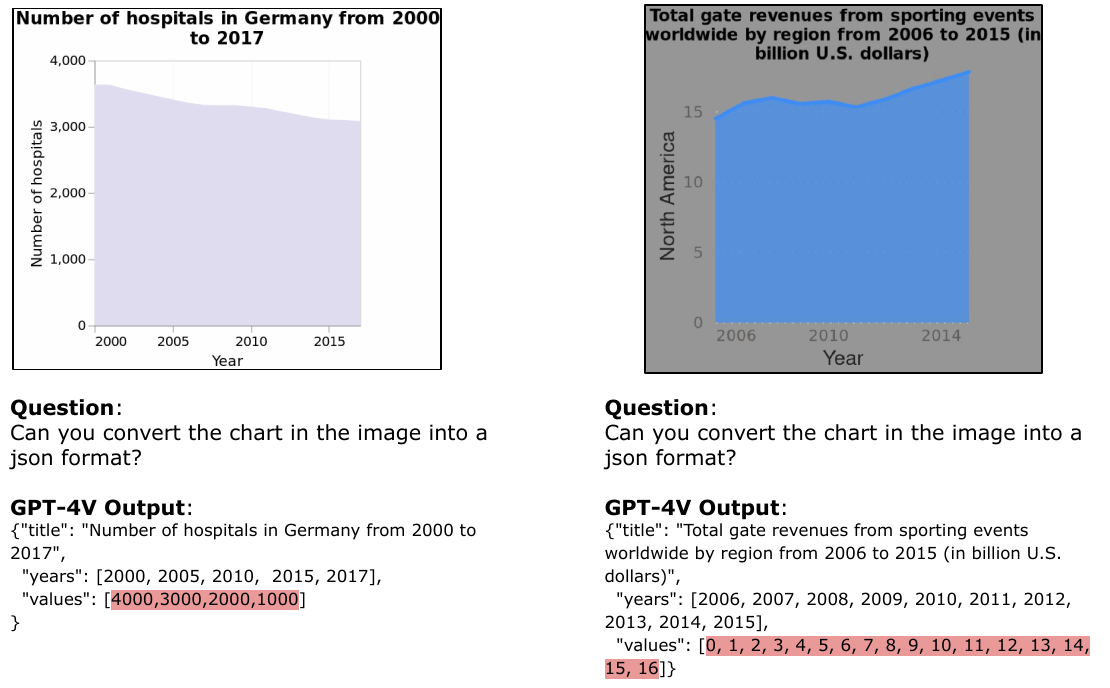}
    \caption{Failure cases of GPT-4V on \textit{Chart to Json} task. \textcolor{red}{RED} means the incorrect answers from GPT-4V. \textbf{Error category}: \textcolor{red}{\textbf{\textit{Vision Perception Error}}}.}
    \label{fig:gpt4_table}
\end{figure*}

\begin{figure*}[h]
    \centering
      \includegraphics[width=0.9\textwidth]{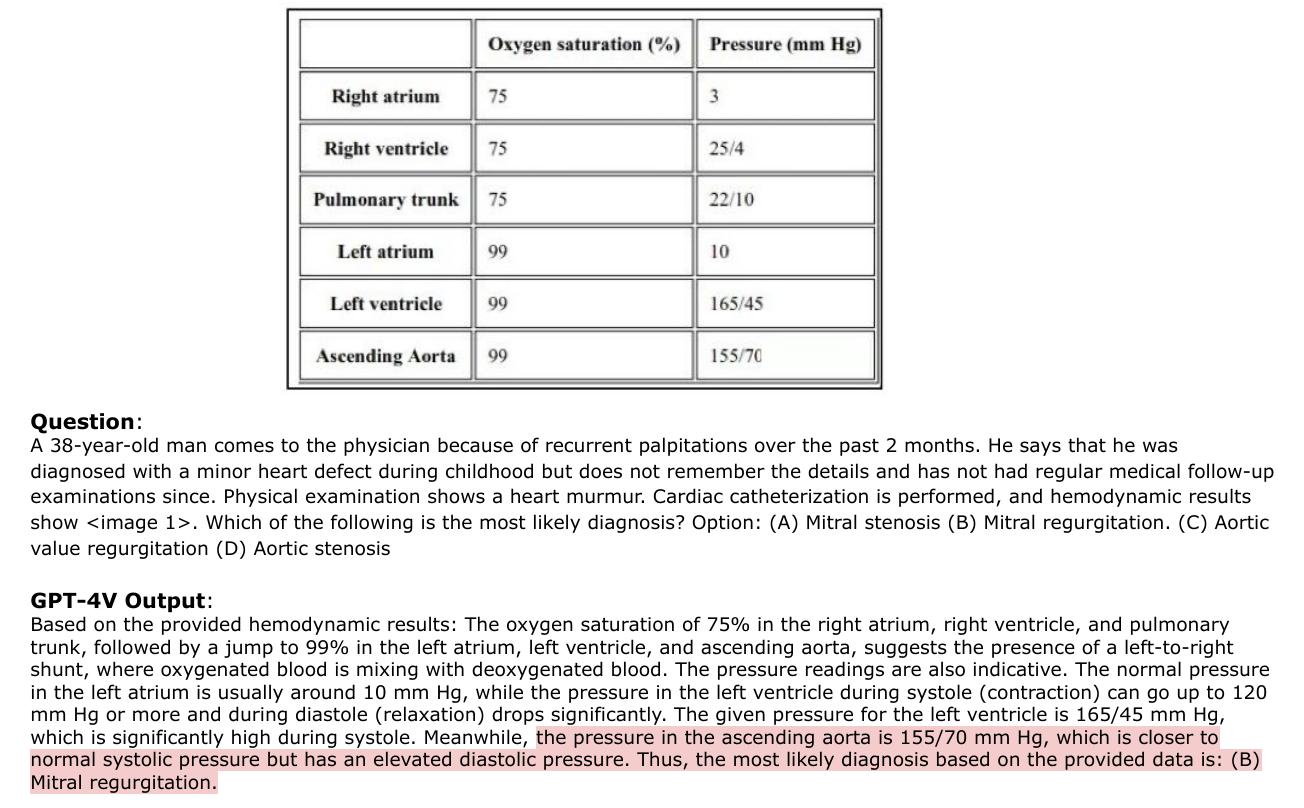}
    \caption{Failure cases of GPT-4V on \textit{Chart to Json} task. \textcolor{red}{RED} means the incorrect answers from GPT-4V. \textbf{Error category}: \textcolor{red}{\textbf{\textit{Lack of Knowledge}}}.}
    \label{fig:gpt4_lack}
\end{figure*}

\begin{figure*}[h]
    \centering
      \includegraphics[width=0.9\textwidth]{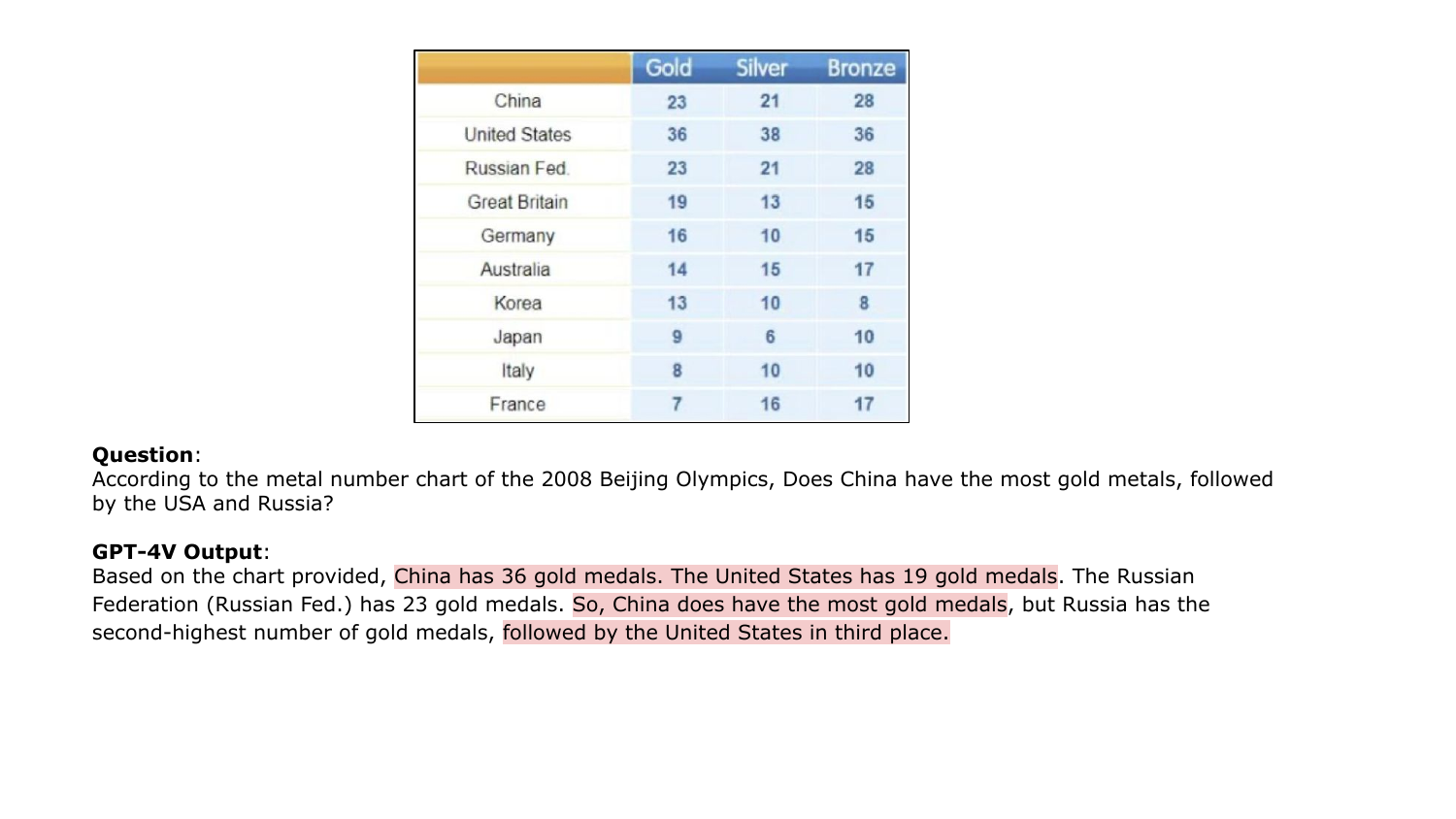}
    \caption{Failure cases of GPT-4V on \textit{Chart to Json} task. \textcolor{red}{RED} means the incorrect answers from GPT-4V. \textbf{Error category}: \textcolor{red}{\textbf{\textit{Vision Perception Error}}} and \textcolor{red}{\textbf{\textit{Language Bias Error}}}.}
    \label{fig:gpt4_mix}
\end{figure*}

\textbf{Other Errors.} The remaining errors include \textbf{Reasoning Error (15\%)} in Fig.~\ref{fig:gpt4_multiple2} and \textbf{Lack of Knowledge (11\%)} in Fig.~\ref{fig:gpt4_lack}. These errors are attributed to various factors, such as complex text interpretation challenges, lack of domain-specific knowledge, or failure to extract precise answers from long context. More cases are shown in Fig.~\ref{fig:gpt4_table} and Fig.~\ref{fig:gpt4_mix}.

\begin{figure}[t]
    \centering
      \includegraphics[width=0.45\textwidth]{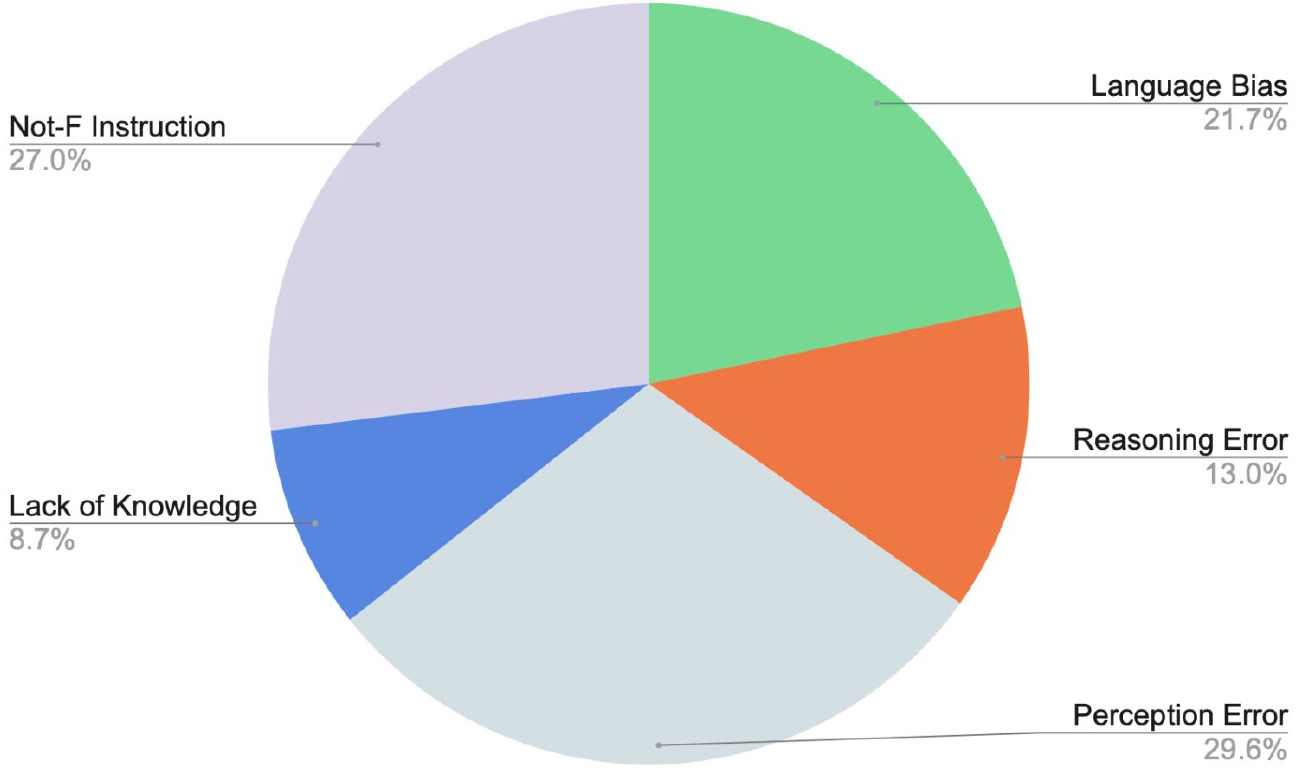}
    \caption{Error distribution of open-source models over 100 randomly sampled error instances. \textbf{\textit{Not-F Instruction}} means \textit{"Not Following Instructions"}.}
    \label{fig:error_open}
\end{figure}

\subsubsection{Error Analysis of Open-Source Models}
\begin{figure*}
     \centering
     \begin{subfigure}[b]{\textwidth}
         \centering
         \includegraphics[width=0.9\textwidth]{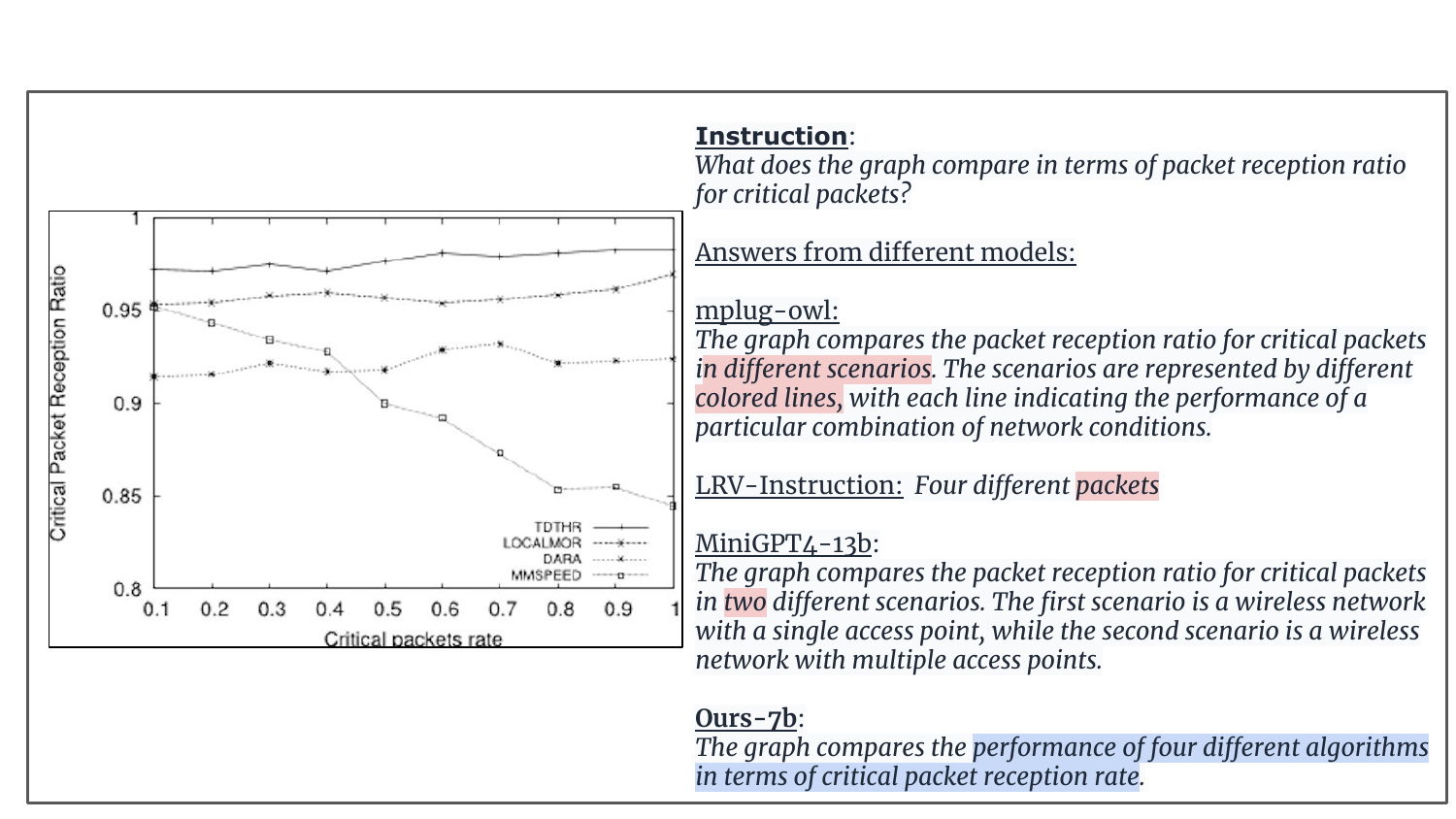}
     \end{subfigure}
     \caption{Examples of \textit{Scientific Chart Understanding} task results from our model and other open-source models in \textit{MMC-Benchmark}. \textcolor{red}{RED} means incorrect answers, and \textcolor{blue}{BLUE} means correct answers.}
     \label{fig:demo14}
\end{figure*}

We examine 100 randomly sampled error instances from open-source models. The instances are analyzed by expert annotators who identify the root causes. The distribution of errors is in Fig.~\ref{fig:error_open}. Different from GPT-4V, one key issue of the open-source model is \textbf{\textit{Not Following Instructions (27\%)}}. Even with a very concise instruction design, there are LMMs that do not follow the user's instructions. For example, in Fig.~\ref{fig:demo12}, when asked \textit{``Please identify the proportion of Americans who favor the coal mining.''}, PixsStruct and MiniGPT-v2 answer \textit{``Yes''} and \textit{``Most Americans favor exporting or expanding solar and wind powers.''}, respectively. In our opinion, a good chart understanding model should be able to follow instructions. However, to the best of our knowledge, most of the existing LLM-based or LMM-based models, except for GPT-4V, are not able to follow human instructions well. More examples are shown in Fig.~\ref{fig:demo11}, \ref{fig:demo13}, and \ref{fig:demo14}.

Another key issue is \textbf{\textit{Vision Encoder is Weak (29.6\%)}}. Existing LMMs typically use CLIP as the vision encoder and do not update its parameters during training. However, as CLIP is trained to align visual embeddings with short captions, its capability of modeling the spatial interactions of chart elements like trend lines and color-coded legends is limited. One potential method is to add segmentation~\cite{kirillov2023segment} and project the segments into the LLM token embedding space. Instead, in our proposed \textit{MMCA} approach, we finetune LMMs on our \textit{MMC-Instruction} data by updating the vision parts during training and improving the integration of visual elements into the LLM input domain. As shown in Tab.~\ref{tab:vision_encoder}, the model without fine-tuning the vision encoder under-performs our proposed MMCA model. It indicates that fine-tuning the vision encoder part of the model is necessary. The improvements in our experiments also demonstrate the effectiveness of our proposed \textit{MMC-Instruction} dataset and the training strategy in \textit{MMCA}. Please refer to Fig.~\ref{fig:demo11}, Fig.~\ref{fig:demo12}, Fig.~\ref{fig:demo13}, and Fig.~\ref{fig:demo14} for more examples.

\subsubsection{More Discussions}
\begin{figure*}
     \centering
     \begin{subfigure}[b]{\textwidth}
         \centering
         \includegraphics[width=0.9\textwidth]{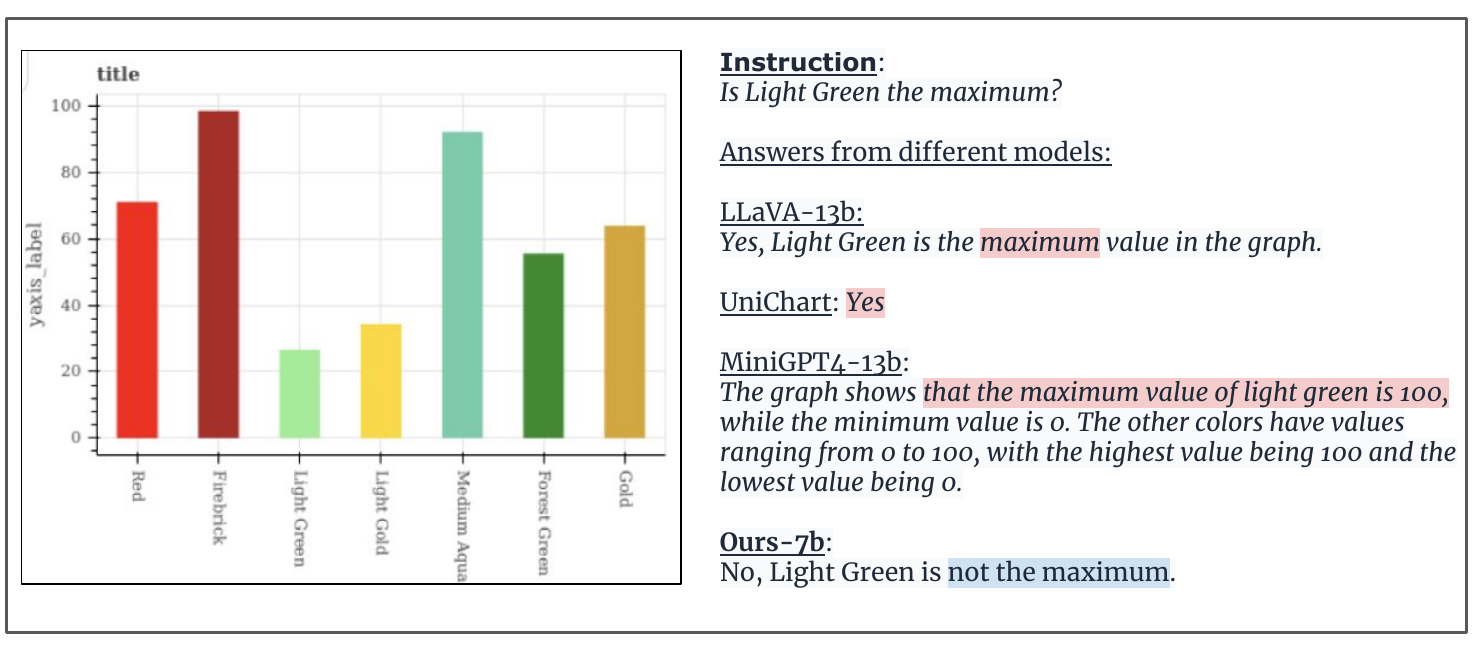}
         \caption{Examples of \textit{Chart Reasoning} task results from our model and other open-source models.}
         \label{fig:demo11}
     \end{subfigure}
     \par\bigskip
     \begin{subfigure}[b]{\textwidth}
         \centering
         \includegraphics[width=0.9\textwidth]{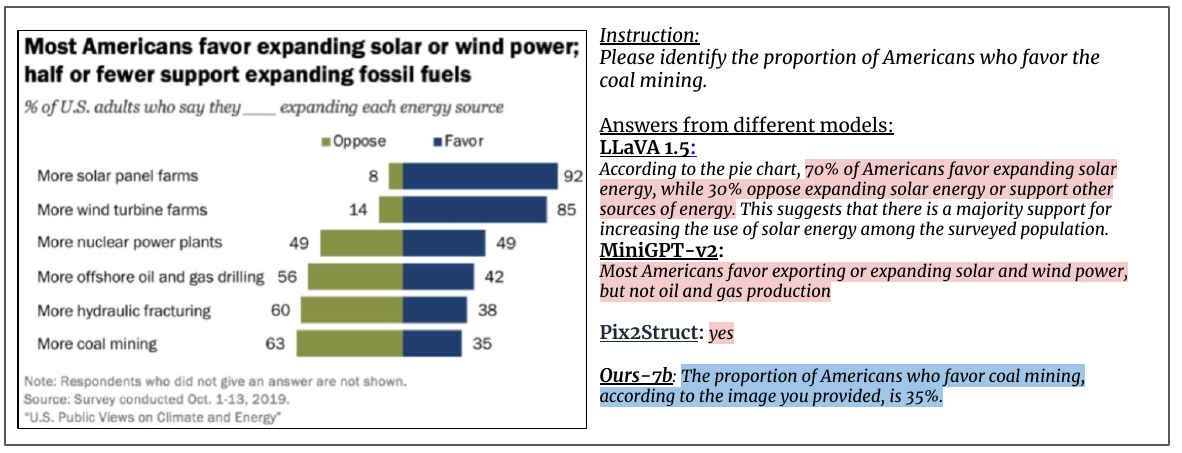}
         \caption{Examples of \textit{Chart Reasoning} task results from our model and other open-source models.}
         \label{fig:demo12}
     \end{subfigure}
     \par\bigskip
     \begin{subfigure}[b]{\textwidth}
         \centering
         \includegraphics[width=0.9\textwidth]{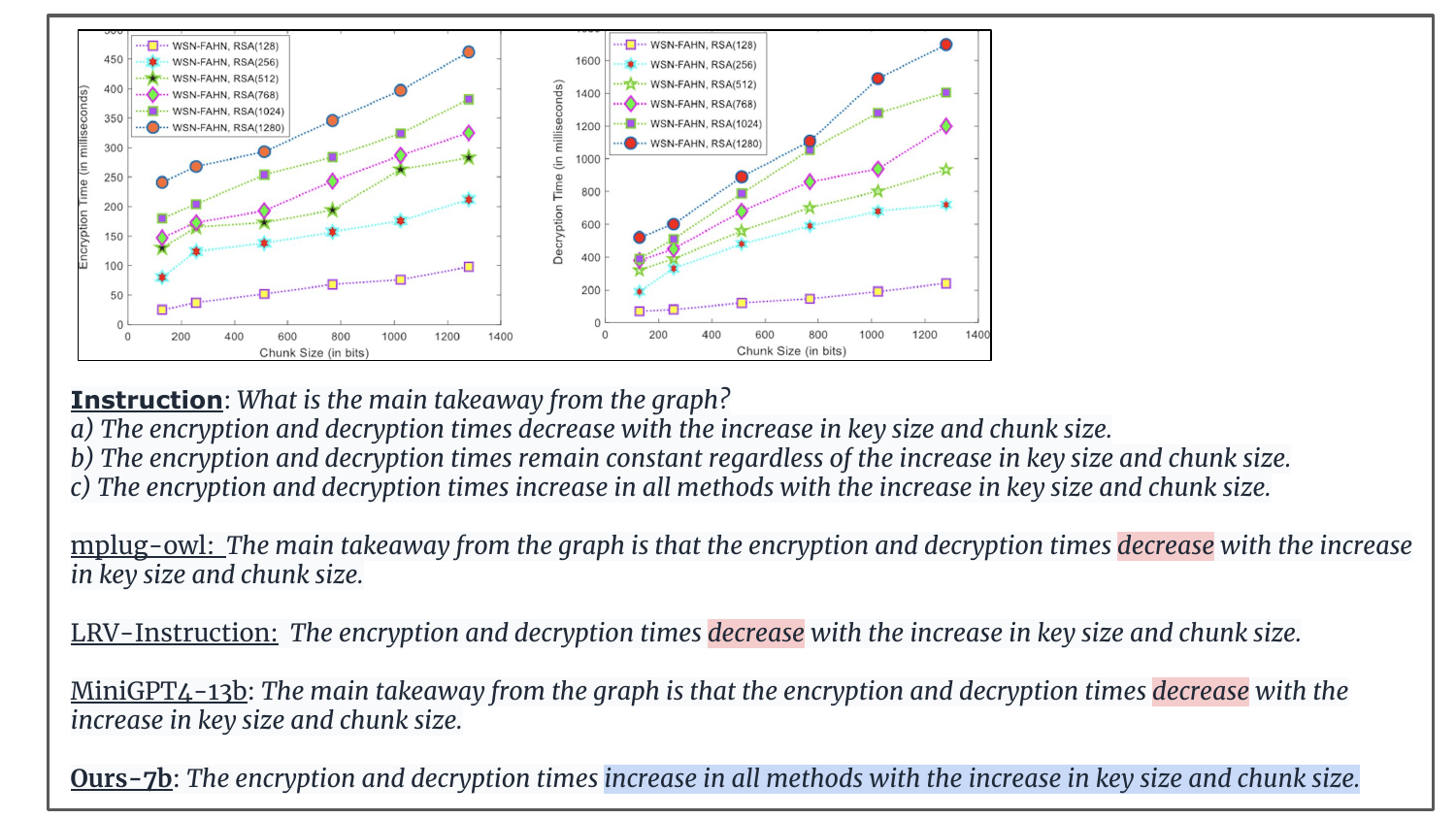}
         \caption{Examples of \textit{Multiple Chart understanding} task results from our model and other open-source models.}
         \label{fig:demo13}
     \end{subfigure}
        \caption{Result examples of our model and other open-source models for three types tasks in \textit{MMC-Benchmark}. \textcolor{red}{RED} means incorrect answers, and \textcolor{blue}{BLUE} means correct answers.}
        \label{fig:demo10}
\end{figure*}

\textbf{\textit{Chart-to-DataTable and Chart-to-Json are extremely Difficult.}} As shown in Tab.~\ref{tab:comparison_mmc}, all current LMMs, including \textbf{\textit{GPT-4V}}, perform badly on these two tasks. It is probably due to the fact that these two tasks require strong OCR skills to output all the data values in the chart correctly. If one value is missing, the prediction will be regarded as incorrect. Compared to the baselines in Fig.~\ref{fig:demo11}, our \textit{MMCA} model is able to produce more accurate responses in correct output formats.

\textbf{\textit{MMC-Benchmark is more Challenging than Previous Benchmarks.}} From Tab.~\ref{tab:comparison_mmc_mvqa}, we find that the overall scores for existing models on \textit{MMC-Benchmark} are lower than those on the current benchmarks like ChartQA. Such results are expected since the questions in \textit{MMC-Benchmark} are more diverse, and the answers are open-ended. Additionally, \textit{MMC-Benchmark} contains more topics that require both a comprehensive understanding of charts and proficient language skills.

\begin{table}[t]
\setlength\tabcolsep{4.3pt}
\centering
\small
\begin{tabular}{lccc}
\toprule[1pt]
\textbf{Model} & ChartQA  & DocVQA & TextVQA\\
\midrule
\textbf{MMCA (Ours)} &  \textbf{57.4} & \textbf{72.5}&\textbf{59.6} \\
-w/o FT Vision Encoder	&54.2	&67.8	&57.2 \\
\bottomrule[1.5pt]
\end{tabular}
\caption{Ablation experiments without fine-tuning vision encoder in MMCA.}
\label{tab:vision_encoder}
\vspace{-0.2in}
\end{table}

\begin{table*}[t]
\setlength\tabcolsep{4.3pt}
\centering
\small
\begin{tabular}{l|c|ccc|c}
\toprule[1pt]
\textbf{MQA Evaluation} & \textbf{Donut} & \textbf{Shikra} & \textbf{BLIP2} & \textbf{InstructBLIP}& \textbf{MMCA (Ours)} \\
\midrule
Chart Information Extraction &  0.46 &  0.38 &  0.36 & 0.41 & \textbf{0.49}\\
Chart Reasoning &  0.42&  0.39 &  0.38 & 0.40 & \textbf{0.47}\\
Contextual Chart Understanding &  0.37 &  0.43 &  0.42 & 0.45 & \textbf{0.55}\\
Multiple Chart Understanding&  0.38 &  0.41 &  0.40 & 0.42 & \textbf{0.47}\\
Chart Type Classification &  0.42 &  0.48 &  0.50 & 0.52 & \textbf{0.59}\\
Chart Topic Classification &  0.45&  0.56 &  0.51 & 0.55 & \textbf{0.64}\\
Stock Chart Analysis &  0.41 &  0.47 &  0.44 & 0.48 & \textbf{0.57}\\
Chart to Datatable &  0.32&  0.39 &  0.40 & 0.41 & \textbf{0.64}\\
Chart to Json  &  0.38&  0.41 &  0.39 & 0.48 & \textbf{0.59}\\
\midrule
Overall &  0.51&  0.47 &  0.42 & 0.45 & \textbf{0.56}\\
\bottomrule[1.5pt]
\end{tabular}
\vspace{0.05in}
\caption{\textit{MMC-Benchmark} evaluation results on Donut, Shikra, BLIP-2, InstructBLIP, and our MMCA regarding the Understanding Ability Evaluation via \textit{Multichoice QA} (MQA) task. We calculate the accuracy of the model predictions in the MQA setting. There is no need to call \textit{GPT-4} for this evaluation.}
\label{tab:comparison_mmc_mvqa_more}
\end{table*}

\begin{figure*}[h]
    \centering
      \includegraphics[width=1\textwidth]{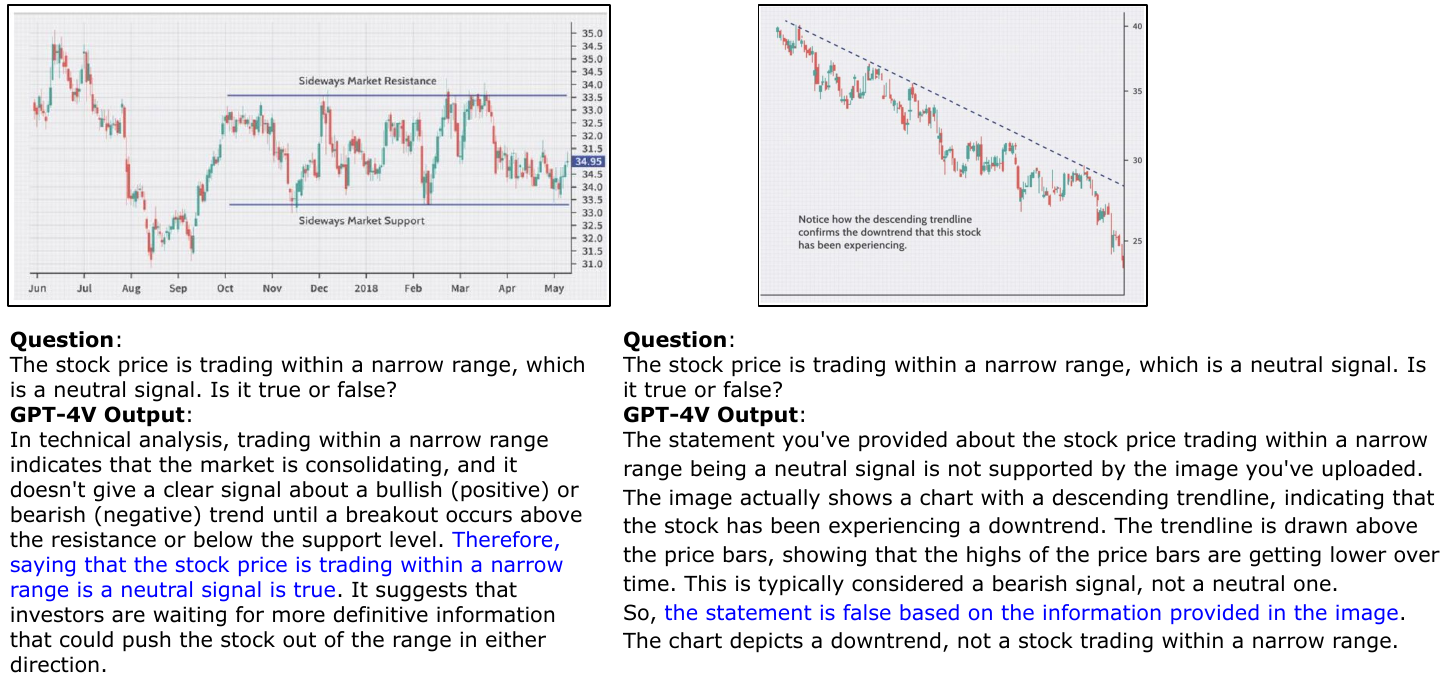}
    \caption{GPT-4V works well on \textit{Stock Chart Analysis} task.}
    \label{fig:gpt4_stock}
\end{figure*}

\begin{figure*}[h]
    \centering
      \includegraphics[width=1.0\textwidth]{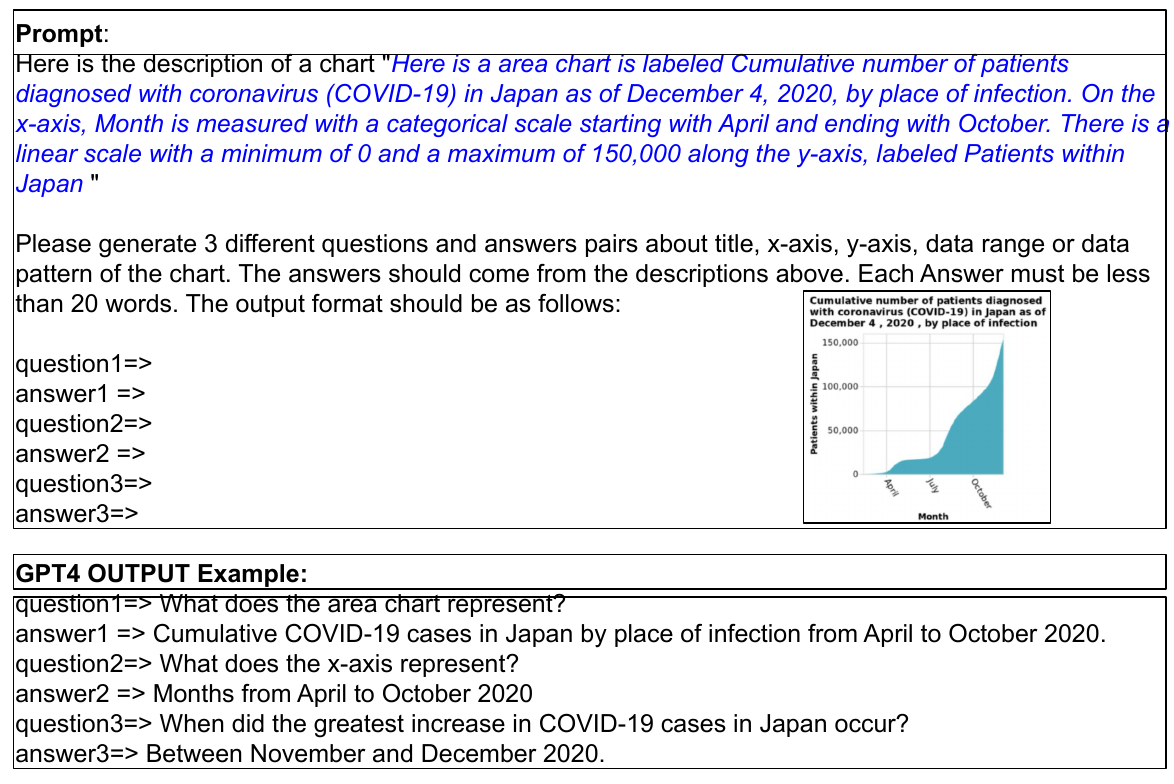}
    \caption{An example prompt for text-only GPT-4 we use to generate instruction and answers for \textit{\textcolor{blue}{Chart Information Extraction}} task. The sentence in \textcolor{blue}{BLUE} is the captions of the chart.}
    \label{fig:chart_prompt_ie}
    \vspace{-0.2in}
\end{figure*}

\begin{figure*}[h]
    \centering
      \includegraphics[width=1.0\textwidth]{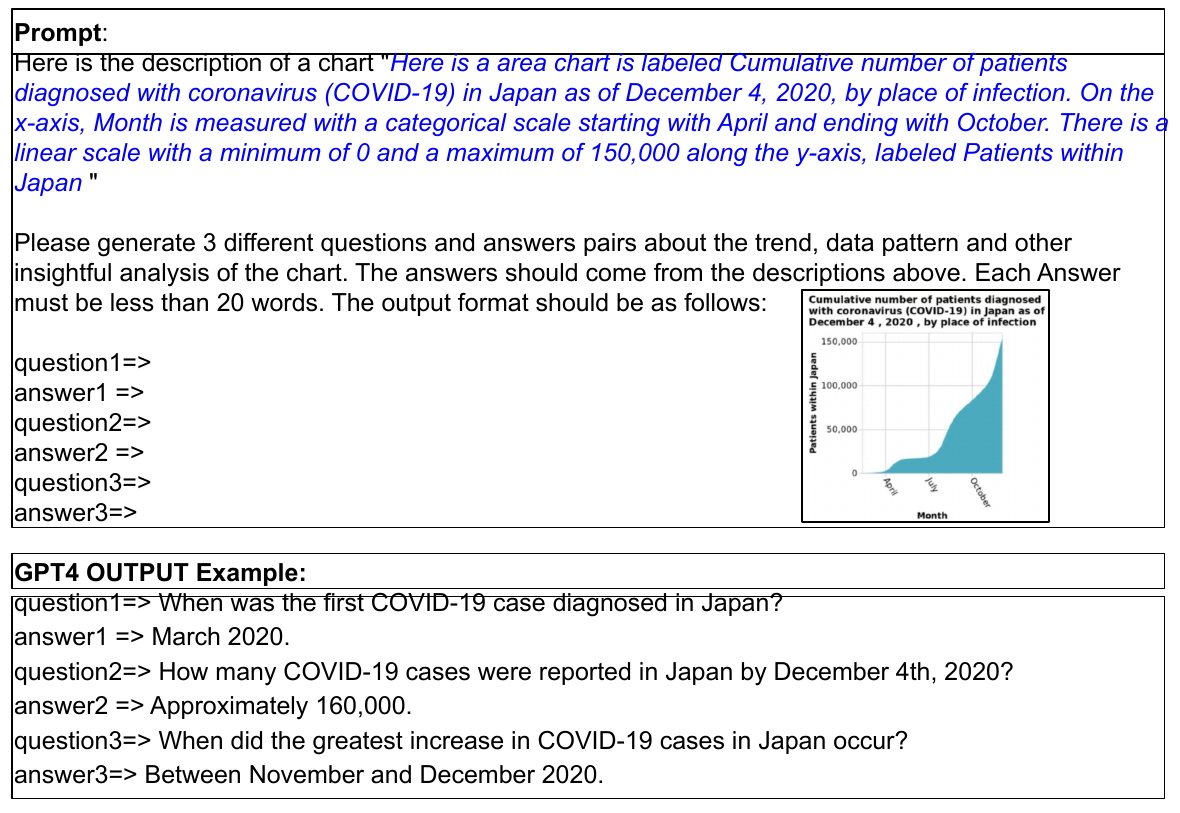}
    \caption{An example prompt for text-only GPT-4 we use to generate instruction and answers for \textit{\textcolor{blue}{Chart Reasoning}} task. The sentence in \textcolor{blue}{BLUE} is the captions of the chart.}
    \label{fig:chart_prompt_cr}
    \vspace{-0.2in}
\end{figure*}

\begin{figure*}[h]
    \centering
      \includegraphics[width=1.0\textwidth]{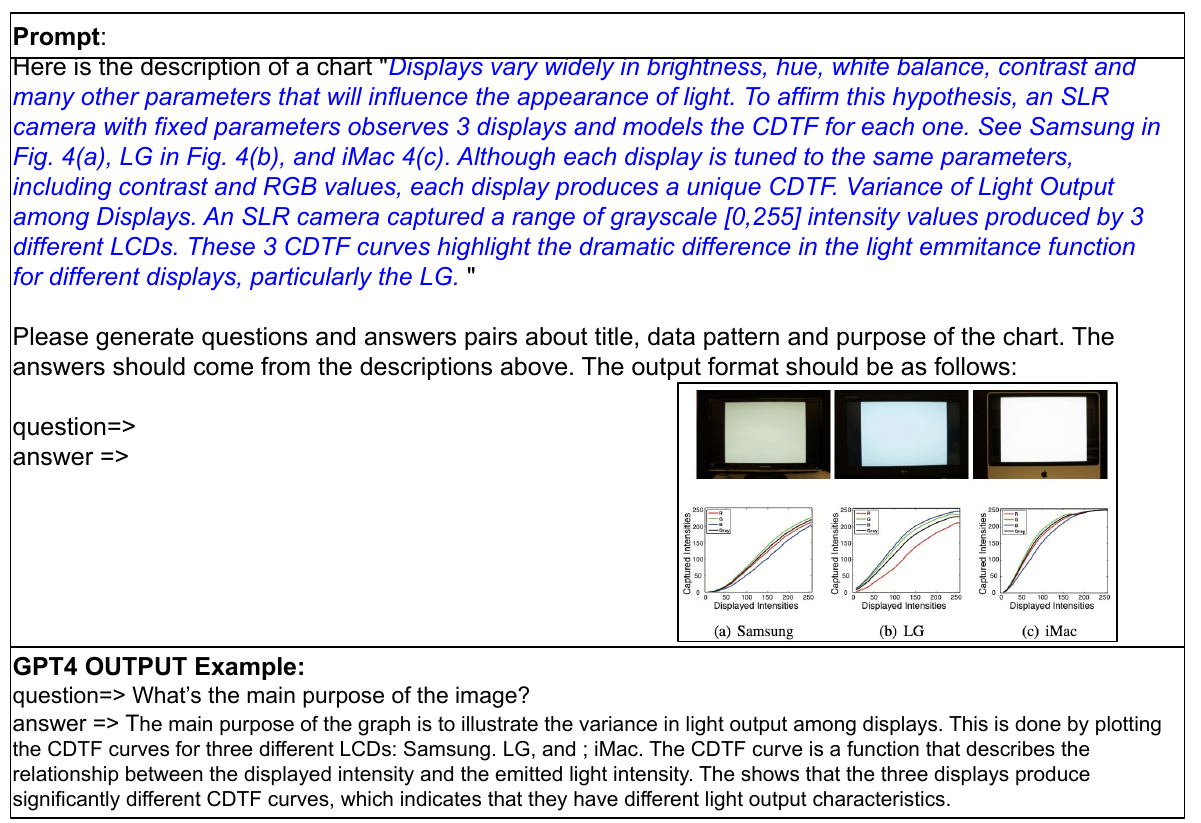}
    \caption{An example prompt for text-only GPT-4 we use to generate instruction and answers for \textit{\textcolor{blue}{Multiple Chart Understanding}} task. The sentence in \textcolor{blue}{BLUE} is the captions of the chart.}
    \label{fig:chart_prompt_arxiv}
    \vspace{-0.2in}
\end{figure*}

\begin{figure*}[h]
    \centering
      \includegraphics[width=1.0\textwidth]{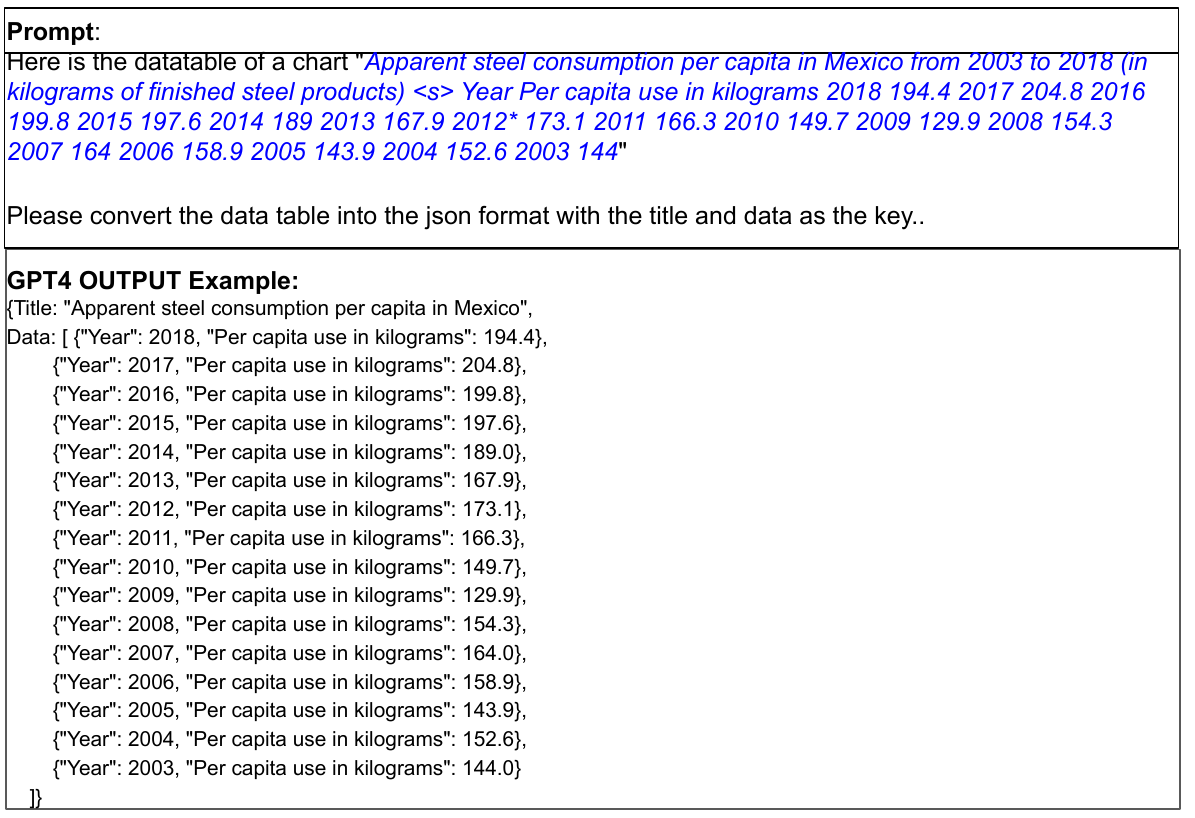}
    \caption{An example prompt for text-only GPT-4 we use to generate instruction and answers for \textit{\textcolor{blue}{Chart-to-Json}} task. The sentence in \textcolor{blue}{BLUE} is the captions of the chart.}
    \label{fig:chart_prompt_json}
    \vspace{-0.2in}
\end{figure*}

\begin{table*}[t]
\setlength\tabcolsep{4.3pt}
\centering
\small
\begin{tabular}{l|cc}
\toprule[1pt]
\textbf{Method} & \textbf{Vision Encoder} & \textbf{Language Model} \\
\midrule
Donut &  ViT-g (1.3B) &  Bert (0.34B)\\
Pix2Struct &  ViT-g (1.3B) &  BART (1.3B)\\
BLIP-2 &  ViT-g (1.3B) &  Vicuna (7B)\\
MiniGPT-v2 &  ViT-g (1.3B) &  Vicuna (7B)\\
LLaVA1.5 &  ViT-L (0.3B) &  Vicuna (7B)\\
mPLUG-Owl &  ViT-L (0.3B) &  LLaMA (7B)\\
InstructBLIP&  ViT-g (1.3B) &  Vicuna (7B)\\
LRV-Instruction &  ViT-L (0.3B) &  Vicuna (7B)\\
\midrule
\textbf{MMCA (Ours)} &  ViT-L (0.3B) &  Vicuna (7B)\\
\bottomrule[1.5pt]
\end{tabular}
\vspace{0.05in}
\caption{The backbones for the vision encoder and language model of the baselines and our MMCA model.}
\label{tab:baseline_size}
\end{table*}
\label{sec:appendix}

\end{document}